\pdfoutput=1

\documentclass[journal, final]{IEEEtran}

\usepackage{float}  

\usepackage{blindtext}
\usepackage{amsmath}

\newcommand{\isep}{\mathrel{{.}\,{.}}\nobreak}

\newcommand{\mat}[1]{\ensuremath{\mathbf{#1}}}
\renewcommand{\vec}[1]{\ensuremath{\mathbf{#1}}}
\newcommand{\vf}[1]{\ensuremath{\mathbf{#1}}}
\newcommand{\Seq}{S} 
\newcommand{\Prior}{P_{\text{pr}}}
\newcommand{\Go}{\text{Go}}  
\newcommand{\Iseq}{I}  
\newcommand{\Normp}[1]{\left\langle {#1} \right\rangle_1} 

\DeclareMathOperator*{\argmin}{argmin}
\DeclareMathOperator*{\argmax}{argmax}
\DeclareMathOperator*{\dist}{dist}
\DeclareMathOperator{\elmult}{\otimes}
\DeclareMathOperator{\eldiv}{\varnothing}
\newcommand{\vecg}[1]{\ensuremath{\boldsymbol{#1}}} 

\DeclareMathOperator{\eff}{eff}
\DeclareMathOperator{\pre}{pre}

\usepackage[]{graphicx}
\graphicspath{{./img/}}

\usepackage{cite}

\usepackage{apptools}
\usepackage{comment}

\usepackage{subfigure}

\usepackage[normalem]{ulem}

\usepackage{xcolor}
\usepackage{amssymb, amsmath}

\usepackage[ruled,linesnumbered]{algorithm2e}

\usepackage[utf8]{inputenc}
\usepackage[T1]{fontenc}

\usepackage{hyperref}
\hypersetup{
    unicode=false,          
    pdftoolbar=true,        
    pdfmenubar=true,        
    pdffitwindow=false,     
    pdfstartview={FitH},    
    pdftitle={Toy Architecture},    
    pdfauthor={Author},     
    pdfsubject={Subject},   
    pdfcreator={Creator},   
    pdfproducer={Producer}, 
    pdfkeywords={keyword1, key2, key3}, 
    pdfnewwindow=true,      
    colorlinks=false,       
    linkcolor=red,          
    citecolor=green,        
    filecolor=magenta,      
    urlcolor=cyan           
}

\usepackage[capitalize]{cleveref}
\crefname{subappendix}{\IfAppendix{section}{appendix}}{\IfAppendix{sections}{appendices}s}

\newcommand{\qq}[1]{``{#1}''}

\IEEEoverridecommandlockouts

\begin{document}

\title{ToyArchitecture: Unsupervised Learning of Interpretable Models of the World}


\author{
    \IEEEauthorblockN{Jaroslav~Vítků\IEEEauthorrefmark{1}, 
    Petr~Dluhoš\IEEEauthorrefmark{1}, 
    Joseph~Davidson, 
    Matěj~Nikl,
    Simon~Andersson,
    Přemysl~Paška, 
    Jan~Šinkora,
    Petr~Hlubuček,
    Martin~Stránský,
    Martin~Hyben,
    Martin~Poliak,
    Jan~Feyereisl,
    Marek~Rosa}
    \IEEEauthorblockA{\textit{GoodAI}
    \\\{name.surname\}@goodai.com}
\thanks{We thank Will Millership for his careful reading and thoughtful suggestions for the manuscript.}
\thanks{\IEEEauthorrefmark{1} Both authors contributed equally to this work.}
}

\maketitle

\begin{abstract}

Research in Artificial Intelligence (AI) has focused mostly on two extremes: either on small improvements in narrow AI domains, or on universal theoretical frameworks which are usually uncomputable, incompatible with theories of biological intelligence, or lack practical implementations. The goal of this work is to combine the main advantages of the two: to follow a big picture view, while providing a particular theory and its implementation. In contrast with purely theoretical approaches, the resulting architecture should be usable in realistic settings, but also form the core of a framework containing all the basic mechanisms, into which it should be easier to integrate additional required functionality.

In this paper, we present a novel, purposely simple, and interpretable hierarchical architecture which combines multiple different mechanisms into one system: unsupervised learning of a model of the world, learning the influence of one's own actions on the world, model-based reinforcement learning, hierarchical planning and plan execution, and symbolic/sub-symbolic integration in general. The learned model is stored in the form of hierarchical representations with the following properties: 1) they are increasingly more abstract, but can retain details when needed, and 2) they are easy to manipulate in their local and symbolic-like form, thus also allowing one to observe the learning process at each level of abstraction. On all levels of the system, the representation of the data can be interpreted in both a symbolic and a sub-symbolic manner. This enables the architecture to learn efficiently using sub-symbolic methods and to employ symbolic inference.

\end{abstract}



%
\IEEEpeerreviewmaketitle

\section{Motivation}\label{s:motivation}

Despite the fact that strong AI capable of handling a diverse set of human-level tasks was envisioned decades ago, and there has been significant progress in developing AI for narrow tasks, we are still far away from having a single system which would be able to learn with efficiency and generality comparable to human beings or animals. While practical research has focused mostly on small improvements in narrow AI domains, research in the area of Artificial General Intelligence (AGI) has tended to focus on frameworks of truly general theories, like AIXI~\cite{Hutter2010}, Causal Entropic Forces~\cite{Wissner-Gross}, or PowerPlay~\cite{Schmidhuber2013}. These are usually uncomputable, incompatible with theories of biological intelligence, and/or lack practical implementations.

Another class of algorithm that can be mentioned encompasses systems that are usually somewhere on the edge of cognitive architectures and adaptive general problem-solving systems. Examples of such systems are: the Non-Axiomatic Reasoning System \cite{Wang}, Growing Recursive Self-Improvers \cite{B2012}, recursive data compression architecture \cite{Franz2015}, OpenCog \cite{Hart2008OpenCogAS}, Never-Ending Language Learning \cite{Carlson}, Ikon Flux \cite{Nivel2007}, MicroPsi \cite{Bach2003}, Lida \cite{Franklin2006} and many others \cite{Kotseruba2017}. These systems usually have a fixed structure with adaptive parts and are in some cases able to learn from real-world data. There is often a trade-off between scalability and domain specificity, therefore they are usually outperformed by deep learning systems, which are general and highly scalable given enough data, and therefore increasingly more applicable to real-world problems.

Finally, at the end of this spectrum there are theoretical roadmaps that are envisioning promising future directions of research. These usually suggest combining deep learning with additional structures enabling, for example, more sample-efficient learning, more human-like reasoning, and other attributes \cite{Mikolov2015, Lake}.

Our approach could be framed as something between the ones described above. It is an attempt to propose a reasonably unified AI architecture\footnote{The term \qq{architecture} is to be taken to mean an autonomous learning and decision system which controls an agent in a virtual/real environment.} which takes into account the big picture, and states the required properties right from the beginning as design constraints (as in~\cite{Bengio2017}), is interpretable, and yet there is a simple mapping to deep learning systems if necessary.

In this paper, we present an initial version of the theory (and its proof-of-concept implementation) defining a unified architecture which should fill the aforementioned gap. Namely, the goals are to:

\begin{itemize}

\item Provide a hierarchical and decentralized architecture capable of robust learning and inference across a variety of tasks with noisy and partially-observable data.

\item Produce one simple architecture which either solves, or has the potential to solve as many of the requirements for general intelligence as possible\footnote{That is according to the holistic design principles of~\cite{Mikolov2015,Nivel2013}.}.

\item Emphasize simplicity and interpretability and avoid premature optimization, so that problems and their solutions become easier to identify. Thus the name \textbf{\textit{“ToyArchitecture”}}.

\end{itemize}

This paper is structured as follows: first, we state the basic premises for a situated intelligent agent and review the important areas in which current Deep Learning (DL) methods do not perform well (\cref{s:modelProperties}). Next, in \cref{s:environment}, we describe the properties of the class of environments in which the agent should be able to act. We try to place restrictions on those environments such that we make the problem practically solvable but do not rule out the realistic real world environments we are interested in. \cref{s:designRequirements} then transforms the expected properties of the environments into design requirements on the architecture. In \cref{s:description} the functionality of the prototype architecture is explained with reference to the required properties and the formal definition in the Appendix. \cref{s:experiments}, presents some basic experiments on which the theoretical properties of the architecture are illustrated. Finally, \cref{s:conclusion} compares the ToyArchitecture to existing models of AI, discusses its current limitations, and proposes avenues for future research.

\section{Required Properties of the Agent} \label{s:modelProperties}

This section describes the basic requirements of an autonomous agent situated in a realistic environment, and discusses how they are addressed by current Deep Learning frameworks.

\begin{enumerate}
\item \textbf{Learning:} Most of the information received by an agent during its lifetime comes without any supervision or reward signal. Therefore, the architecture should learn in a primarily unsupervised way, but should support other learning types for the occasions when feedback is supplied.

\item \textbf{Situated cognition:} The architecture should be usable as a learning and decision making system by an agent which is situated in a realistic environment, so it should have abilities such as learning from non-i.i.d. and partially observable data, active learning\cite{Hay2018}, etc.

\item \textbf{Reasoning:} It should also be capable of higher-level cognitive reasoning (such as goal-directed, decentralized planning, zero shot learning, etc.). However, instead of needing to decide when to switch between symbolic/sub-symbolic reasoning, the entire system should hierarchically learn to compress high-dimensional inputs to lower-dimensional (a similar concept to the semantic pointer~\cite{Blouw2016}), slower changing~\cite{Wiskott2002}, and more structured~\cite{Machery} representations. At each level of the hierarchy, the same inference mechanisms should be compatible with both (simple) symbolic and sub-symbolic terms. This refers to one of the most fundamental problems in AI\textemdash chunking: how to efficiently convert raw sensory data into a structured and separate format~\cite{Shastri,Marblestone2016}. The system should be able to learn and store representations of both simple and complex concepts to that they can be efficiently reused.

\item \textbf{Biological inspiration:} The architecture should be loosely biologically plausible~\cite{Marblestone2016,Hassabis2017,Hawkins2006}. This means that principles that are believed to be employed in biological networks are preferred (for example in~\cite{Lillicrap2014}) but not required (as in~\cite{Lazaro-Gredilla2016}). The entire system should be as uniform as possible and employ decentralized reasoning and control~\cite{Eisenreich2016}
\end{enumerate}

Recent progress in DL has greatly advanced the state of AI. It has demonstrated that even extremely complex mappings can be learned by propagating errors through multiple network layers. However, deep networks do not sufficiently address all the requirements stated above. The problems are in particular:
\begin{enumerate}

\item Networks composed of unstructured layers of neurons may be too general; therefore, gradient-based methods have to \qq{reinvent the wheel} from the data for each task, which is very data-inefficient. Furthermore, these gradient-based methods are susceptible to problems such as vanishing gradients when training very deep networks. These drawbacks are partially addressed by transfer learning~\cite{QiangYang2010} and specialized differentiable modules~\cite{Hochreiter1997, Santoroa, Santoro, He2015}.

\item The inability to perform explaining-away efficiently, especially in feedforward networks. This starts to be partially addressed by~\cite{Sabour, George2017a}.

\item Deep networks might form quite different internal representations than humans do. The question is whether (and if so: how?) DL systems form conceptual representations of input data or rather learn surface statistical regularities~\cite{Jo2017}. This could be one of the reasons why it is possible to do various kinds of adversarial attacks~\cite{Szegedy, Su} on these systems.

\item The previous two points suggest that deep networks are not interpretable enough, which may be a hurdle to future progress in their development as well as pose various security risks.

\item The inability to build a model of the world based on a suitable conceptual/localist representation~\cite{Roy2011, Bach, Feldman2013} in an unsupervised way leads to a limited ability to reuse learned knowledge in other tasks. This occurs especially in model-based Reinforcement Learning which, for the purposes of this paper, is more desirable than emulating model-free RL~\cite{Deisenroth2011} owing to its sample efficiency. Solving this problem in general can lead to systems which are capable of gradual (transfer/zero-shot \cite{Higginsb, Ha2018}) learning.

\item Many learning algorithms require the data to be i.i.d., a requirement which is almost never satisfied in realistic environments. The learning algorithm should ideally exploit the temporal dependencies in the data. This has been partially addressed e.g. in~\cite{Mnih2015, Blundell2016}.

\item One of unsolved problems of AI lies in sub-symbolic/symbolic integration~\cite{Besold2017,Kotseruba2017}. Most successful architectures employ either just symbolic or sub-symbolic representations. This naturally leads to the situation that sub-symbolic deep networks which operate with raw data are usually not designed with higher-level cognitive processing in mind (although there are some exceptions~\cite{spauninstructions2013}).

\end{enumerate}

Some of the mentioned problems are addressed in a promising \qq{class} of cortex-inspired networks~\cite{Canziani2017}. But these usually aim just for sensory processing~\cite{Rasmus, Laurent2016, Rinkus2014a, Oreilly2017, Qiu2019}, their ability to do sensory-motoric inference is limited~\cite{Laukien}, or they focus only on sub-parts of the whole model~\cite{Hawkins2015}.

\section{Environment Description and Implications for the Learned Model} \label{s:environment}

In order to create a reasonably efficient agent, it is necessary to encode as much knowledge about the environment as possible into its prior structure\textemdash without loss of universality over the class of desired problems. In other words, we are not aiming for an artificial intelligence which is universally general in all possible hypothetical universes (which might not even be possible~\cite{nflt1997}), but rather for an efficient and multi-purpose machine tailored to a chosen class of environments.

 We consider realistic environments with core properties (such as space and time) following from physical laws. The purpose of this section is to describe the assumed properties of the environment and their implications for the properties of the world model. In the following, the process which determines the environment behavior will be called the \textit{Generator}, while the model of this process learned by the agent will be called the \textit{(Learned) Model}. 

For simplicity, we first consider a passive agent which is unable to follow goals or interact with the environment using actions. In \cref{s:description}, we extend both the Model and the Generator by considering actions and reinforcement signals as well. There are multiple properties which we desire of the Generator.

\subsection{Stationarity}

The dynamics of the environment is generated by a stationary process (or a non-stationary one which is changing slowly enough for the agent to adapt its learned model to the changes).

\subsection{Non-linearity, Continuity and Partial Observability}

Real environments are typically continuous and partially observable. Their Generators can be modeled as general non-linear dynamical systems:
\begin{align} 
    \dot{x} &= f(x,u)+w, \nonumber \\
    o &= g(x,u) + z, \label{e:nlds}
\end{align}
where the state transition function $f$ and observation function $g$ are nonlinear functions taking state variable $x$ and inputs $u$ as parameters, the $\dot{x}$ is the derivative of $x$. The function $f$ changes the state variable, while the function $g$ produces observations $o$ which can be perceived by the agent. The terms $z$ and $w$ denote noise~\cite{Friston2008,Socolar2006}. This means that hidden states are not observed directly; rather, they have to be estimated indirectly from the observations $o$. 

\subsection{Non-determinism and Noise} \label{s:nonDeterministicAndNoisy}

Even though the internal evolution of realistic environments may be deterministic, they are often complex and typically have non-observable hidden states. An observation function $o$ for these environments will thereby impart incomplete information. Additionally, the sensors of the agent are imprecise, and thus there is inherent noise ($z$ in \cref{e:nlds}) so the reading of $o$ (even if it is for a fully observable world) may be flawed. We can model this uncertainty (be it for faulty sensors or non-observability) by expressing the Generator as a stochastic process.

\subsection{Hierarchical Structure and Spatial and Temporal Locality} \label{s:predominantlyHierarchical}

It is reasonable to expect that the agent will interact with an environment that has many hidden state variables and very complex functions for  state-transitions and observations: $f$ and $g$ in \cref{e:nlds}. Learning in this setting is not a tractable task in general. Therefore, we will include additional assumptions based on properties of the real world.

\begin{figure}[ht]
    \centering
    \includegraphics[width=\linewidth]{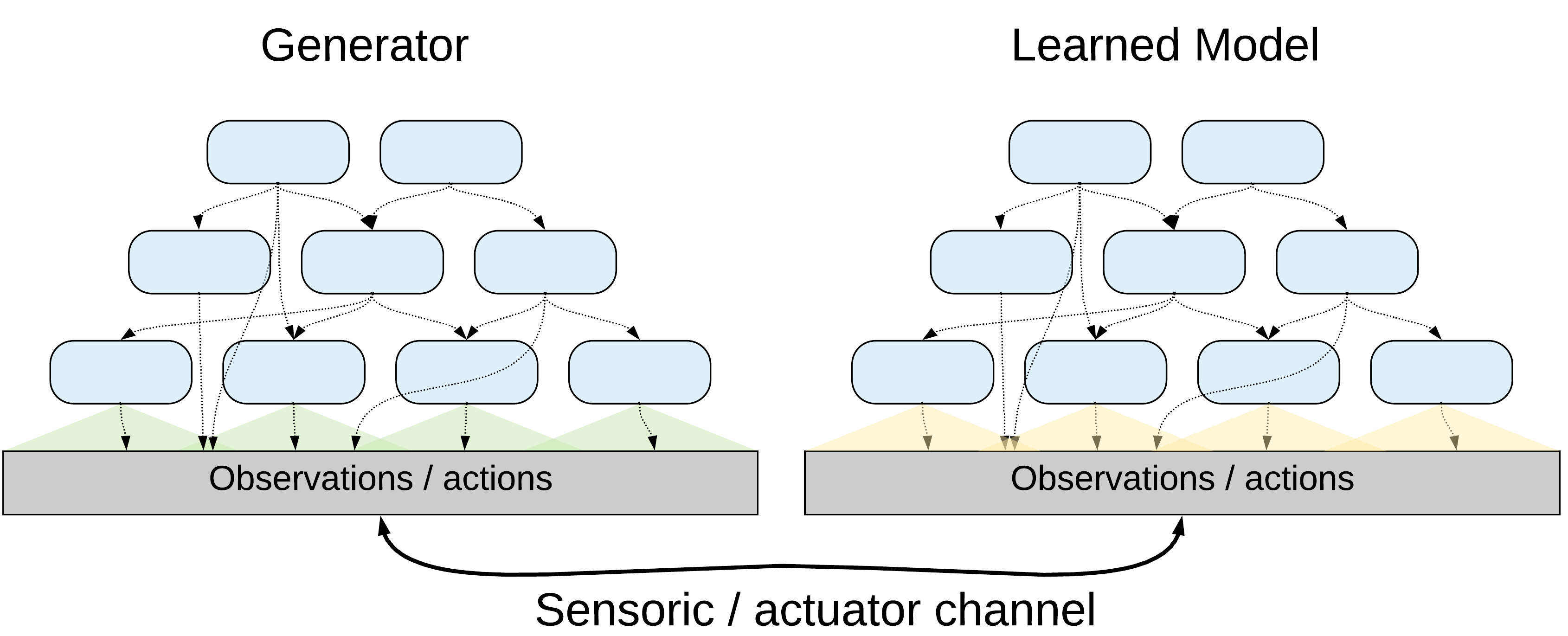}
    
    \caption{The Hierarchical Generator (left), which generates spatially and temporally localized observable patterns. The Learned Model in the agent (right) should ideally correspond to the structure of the Generator. The agent's sensors and actuators are localized in the environment as in~\cite{D1987,Friston2008}. Note that in many cases a single observation is a mix of effects of multiple sub-generators  running in parallel.}
      
    \label{f:hierarchies}
\end{figure}

We assume that the Generator has a predominantly hierarchical structure~\cite{Franz, Lin}, both in space and time; therefore, it can be modeled as Hierarchical Dynamic Model (HDM)~\cite{Friston2008}. We expect that the observations generated by such system are both local in space (one event influences mostly events which share similar spatial locations) and in time (subsequent observations share more information than distant ones), as described by the following power law relations:
\begin{equation}
\label{e:locality}
\begin{aligned}
	I \big (o_i(t); o_j(t) \big ) 	& \propto \dist(i,j)^{-\text{const}},\\
    I \big (o(t); o(t+\Delta) \big ) 	& \propto \Delta^{-\text{const}},
\end{aligned}
\end{equation}
where $I(x; y)$ is a measure of mutual information between variables $x$ and $y$, $dist()$ is a spatial distance function appropriate for the particular environment (e.g., Euclidean distance between pixels in an image), $\Delta$ is temporal distance, and $const$ is a positive constant. 

Note that both requirements are not strict and allow sporadic non-hierarchical interactions, interactions between small details in spatially/temporally distant events. 

These relations reflect a common property of real world systems\textemdash that they have structure on all scales~\cite{Franz, Lin2017a}. It can serve as an inductive bias enabling the agent to learn models of environments in a much more efficient way by trying to extract information on all levels of abstraction. These assumptions also reveal an important property that data perceived and actions performed by the agent are highly non-i.i.d., which has to be taken into consideration when designing the agent.

Another important property of such a hierarchy is that at the lowest levels, most of the information (objects, or their properties) should be \qq{place-coded} (e.g. by the fact that a  sub-generator on a particular position is active/inactive), but as we ascend the hierarchy towards more abstract levels, the information should be more \qq{rate-coded} in that we keep track of the state of particular sub-generators (e.g. their hidden states or outputs) through time~\cite{Sabour}. This means that in higher levels, the representation should become more structured and local.

\subsection{Decentralization and High Parallelism}
\label{envA:parallel}
The spatial locality of the environment implies that on the bottom of the Generator hierarchy, each of the sub-generators influences a spatially localized part of the environment. In realistic environments it is usually true that multiple things happen at the same time. This implies that a single observation should be a mix of results of multiple sub-generators (relatively independent sub-processes/causes) running in parallel, similar to Layered HMMs~\cite{Oliver2004}.
\section{Design Requirements on the Architecture} \label{s:designRequirements}

The assumptions about the Generator described in the previous section were derived from the physical properties of the real world. They serve as a set of constraints that can be taken into account when designing the architecture to model these realistic environments. Such constraints should make the learning tractable while retaining the universality of the agent. 

The goal is to place emphasis on the big picture and high-level interactions within parts of the architecture while still providing some functional prototype. Therefore, individual parts of the presented architecture are as simple and as interpretable as possible. Many of the implemented abilities share the same mechanisms, which results in a universal yet relatively simple system.

The sensors of any agent situated in a realistic environment have a limited spatial and temporal resolution, so the agent is in reality observing a discrete sequence of observations $O = o_1, o_2, \ldots, o_T$, each drawn from an intractable but finite vocabulary $\mathcal{A} = a_1, a_2, \ldots, a_{|\mathcal{A}|}$. Thus, it could be possible to approximate the Generator by a Hidden Markov Model (HMM) with enough states.

An approximation more suitable for the hierarchical structure of the Generator is the Hierarchical Hidden Markov Model (HHMM)~\cite{Fine1998}. It is a generalization of the HMM where each state is either a production state (a leaf node that emits an observation) or a hidden state which itself represents an HMM. The HMM generates sequences by recursive activation of one of the states in the model (vertical transition) until the production state is encountered. After this, control is handed back to the parent HMM where a horizontal transition is made. The HHMM can be converted into an HMM by concatenating the observation-emitting states and recomputing the transition probabilities. Note that this is a relatively general approach which is similar to Linear Time-Invariant (LTI) dynamical systems~\cite{Roweis1999}.     
However, the HHMM has two substantial limitations, namely the inability to efficiently reflect (rare) non-hierarchical relationships between subparts (two neighboring sub-processes cannot directly share any information about their states) and its serial nature.

In order to efficiently address the fact that the Generator is parallel (and therefore, each observation can contain results of multiple sub-processes mixed together), the architecture has to be able to learn how to disentangle~\cite{Higgins, Bengio2017} independent events from each other, and continue to do so on each level of the learned hierarchy.

We will show that the architecture presented in this paper overcomes both aforementioned limitations of HHMM and can efficiently approximate the Generator described in the previous sections. Namely, it can operate in continuous environments (similar to semi-HMMs \cite{baum1966}), but it can also automatically chunk the continuous input into semi-discrete pieces. It can process multiple concurrently independent sub-processes (an example of this is a multimodal sensor data fusion as in Layered HMMs~\cite{Oliver2004}), and can handle non-linear dynamics of the environment. Finally, the architecture presented here can handle non-hierarchical interactions via top-down or lateral modulatory connections, which are often called the context~\cite{Richert2016,Hawkins2017,Adams,Canziani2017}.

\subsection{Hierarchical Partitioning and Consequences}\label{s:hierarchicalPartitioning}

Due to the fact that the interactions are largely constrained by space and time, the generating process can be seen as mostly decentralized, and it is reasonable to also create the Learned Model as a hierarchical decentralized system consisting of (almost) independent units, which we call \textit{Experts}. In the first layer, each Expert has a spatially limited field of view\textemdash it receives sequences of local subparts of the observations from the Generator (see \cref{f:hierarchies}). The locality assumptions in \cref{e:locality} suggest that such a localized Expert should be able to model a substantial part of the information contained in its inputs without the need for information from distant parts of the hierarchy.

The outputs of Experts in one layer serve as observations for the Experts in subsequent layers, which have also only localized receptive fields but generally cover larger spatial areas, and their models span longer time scales. They try to capture the parts of the information not modelled by the lower layer Experts, in a generally more abstract and high-level form.

Each Expert models a part of the Generator observed through its receptive field using discrete states with linear and serial (as opposed to parallel) dynamics. In an ideal case, the Expert's receptive field would correspond exactly to one of the local HMMs: 
\begin{equation}
\label{e:linearHmm}
\begin{aligned} 
    \vec{s}(t+1)  &= \mat{A} \vec{s}(t),\\
    \vec{o}(t)    &= \mat{B} \vec{s}(t),
\end{aligned}
\end{equation}
 where the $\mat{A}$ is a transition matrix and $\mat{B}$ is an observation emission matrix.

But in reality, one Expert can see observations from multiple neighboring Generator HMMs, it might not see all of the observations and does not know about the sporadic non-hierarchical connections, so the optimal partitioning of the observations and the exact number of states for each Expert is not known a priori and in general cannot be determined. Therefore, the architecture starts as a universal hierarchical topology of Experts and adapts based on the particular data it is observing. Although all the parameters of the topology and the Experts could be made learnable from data (e.g. the number of Experts, their topology, the parameters of each Expert), we decided to fix some of them (e.g. the topology) or set them as hyperparameters (e.g. the parameters of each Expert). Therefore, the current version of the architecture uses the following two assumptions:

\begin{itemize}
\item The local receptive field of each Expert is defined a priori and fixed.
\item The number of hidden states of the model in each Expert is chosen a priori and fixed as well.
\end{itemize}

These assumptions (see \cref{f:approxHierarchy}) have the following implications:

\begin{itemize}
\item An Expert might not perceive all the observations that are necessary to determine the underlying sub-process of the Generator responsible for the observations.
\item An Expert might not have sufficient resources (e.g. number of hidden states/sequences) to capture the underlying sub-process.
\end{itemize}

\begin{figure}[t]
    \includegraphics[width=0.9 \linewidth]{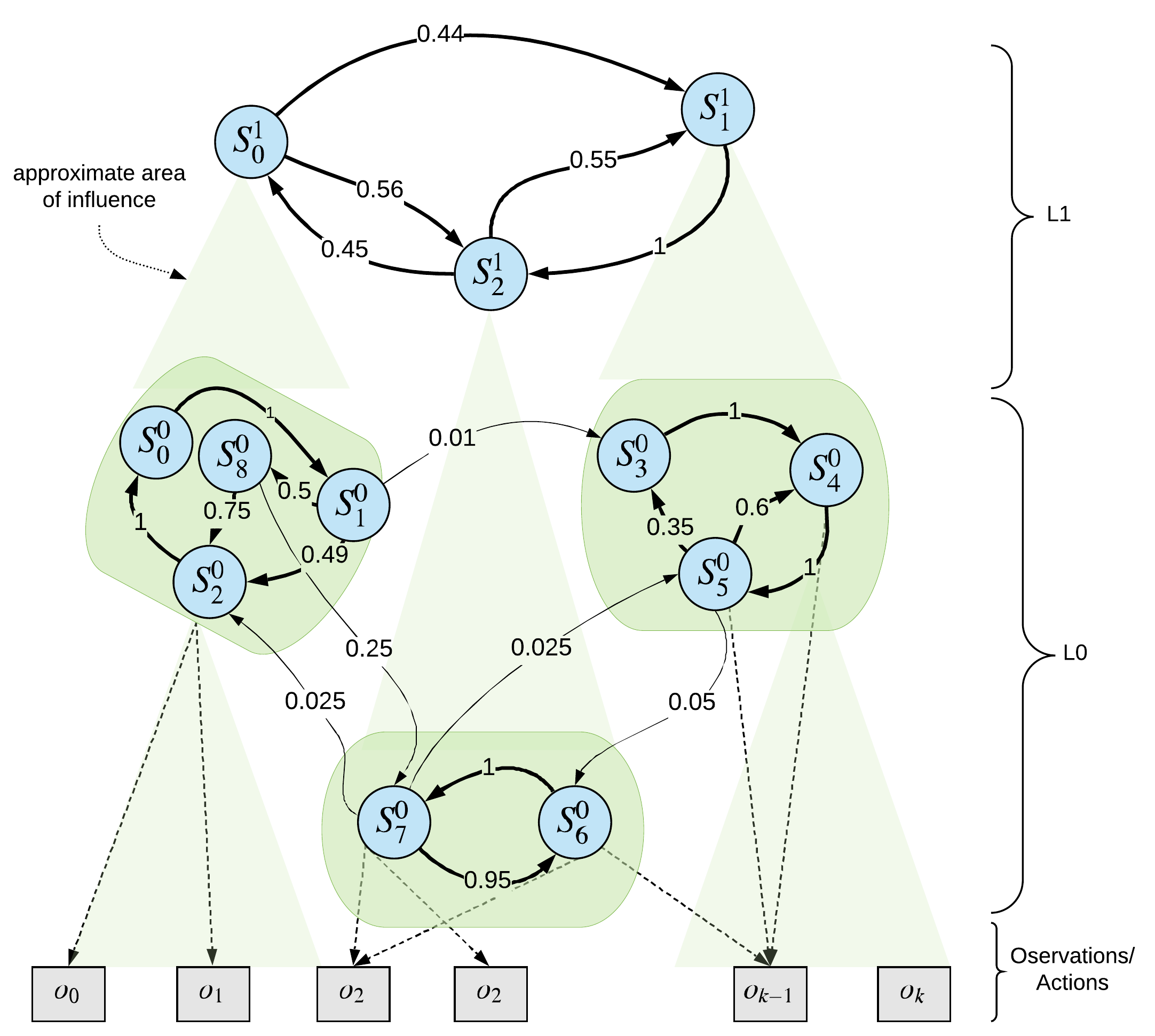}
    \centering
    
    \caption{Example of the hierarchical structure of the world which fulfills the locality in space assumption, and has a fixed number of hidden states. The hierarchy has two levels, in $L0$ there are 3 parallel Markov models, and one in $L1$ on the top. The $S_i^j$ denotes state $i$ in layer $j$, numbers on the edges are illustrative transition probabilities.}
      
    \label{f:generator}
\end{figure}

\begin{figure}[t]
    \includegraphics[width=0.9 \linewidth]{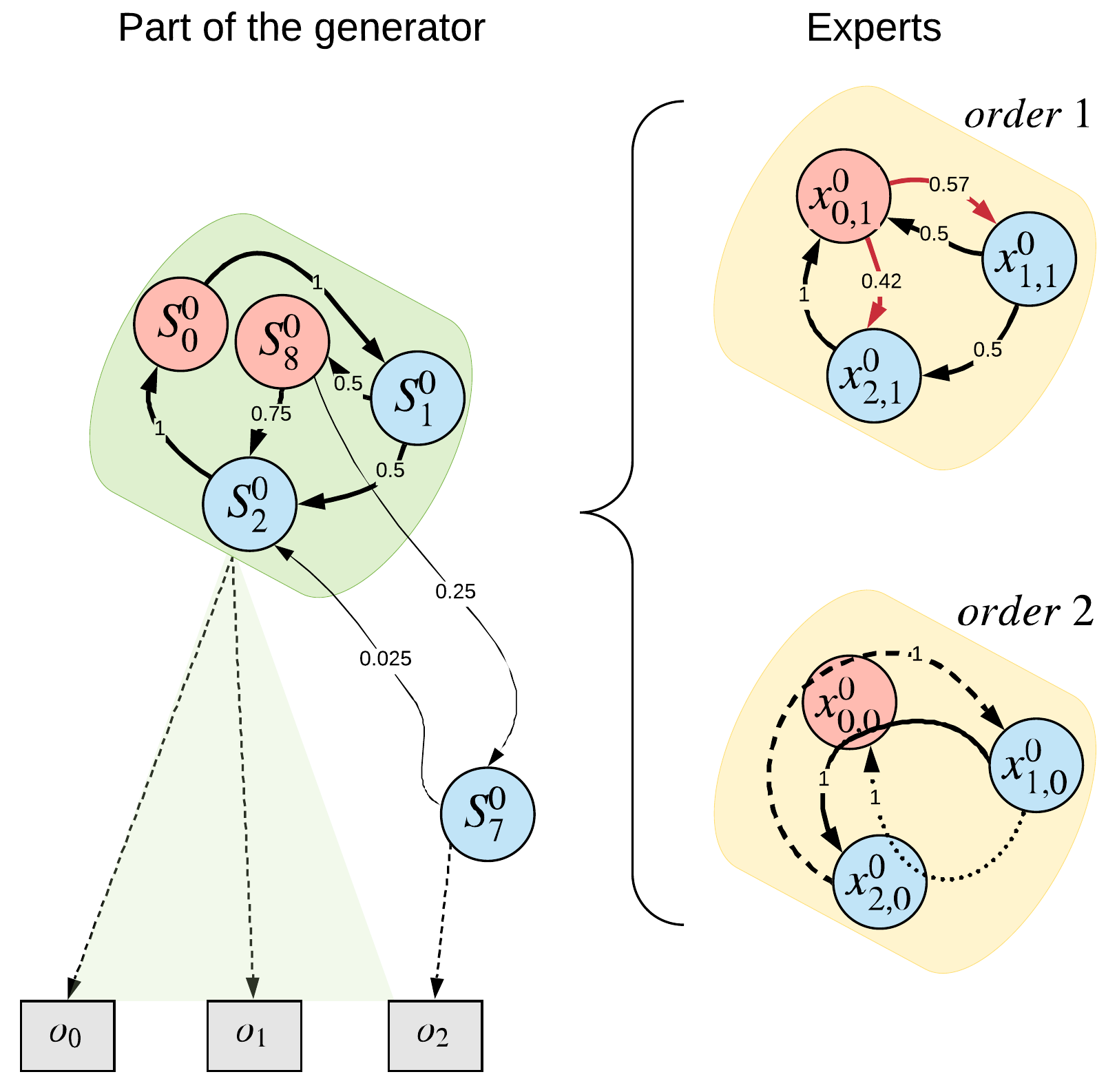}
    \centering
    
    \caption{Approximation of one Markov model from the Generator shown in \cref{f:generator}.  Here, one part of the Generator (\textit{green box}) is approximated by two Experts (\textit{yellow boxes}). While both Experts have insufficient number of states, the bottom one mitigates this problem by increasing the order of its Markov chain (parameter $T_h$ in Eq. \eqref{e:longerHmm}). The $x_{i,k}^j$ denotes the hidden state $i$ of Expert $k$ in layer $j$. The top Expert $k=1$ (which models the process with Markov order 1) shows that the original process cannot be learned well if it has an insufficient number of states (the red Expert states corresponds to the red Generator states $S_0^0$ and $S_8^0$ in the original process). Given the state $x_{0,1}^0$, the Expert is unable to predict the probabilities of the next states correctly.
    Compared to this, the bottom Expert $k=0$ models the process with Markov order 2 ($T_h=1$), therefore the probabilities of the next states depend on the current and previous state (indicated by arrows across 3 states in the image). In this case, despite the fact that $x_{0,0}^0$ is ambiguous, the bottom Expert can correctly predict the next state of the original process (for simplicity, transition probabilities are illustrative and not all are depicted).}
      
    \label{f:approxHierarchy}
\end{figure}

Note that even without the aforementioned assumptions, with the ideal structure and topology of the Experts, their models would not correspond exactly to the Generator until fully learned, which can be impossible to achieve due to limited time and limited information being conveyed via the observations. Therefore, the architecture has to be robust enough so that multiple independent sub-processes of the Generator can be modeled by one Expert, and conversely, multiple Experts might be needed to model one subprocess. Such Experts can then be linked via the context channel (see Appendix \ref{s:externalContext}). It is a topic of further research whether, and how much, fixing each parameter limits the expressivity and efficiency of the model.  

So instead of modelling the input as one HMM as described in Eq. \eqref{e:linearHmm}, each Expert is trying to model the perceived sequences of observations $\vec{o}$ using a predefined number of hidden states $\vec{x}$ and some history of length $T_h$.

Additionally, we define an output projection function computing the output of the Expert $\vec{y}$:
\begin{equation}
\begin{aligned}
\label{e:longerHmm}
\vec{x}(t+1)    & = f_1(\vec{x}(t), \vec{x}(t-1), \ldots, \vec{x}(t-T_h)),\\
\vec{o}(t)  & = f_2(\vec{x}(t)),\\
\vec{y}(t) & = f_3(\vec{x}(t), \vec{x}(t-1), \ldots, \vec{x}(t-T_h)),
\end{aligned}
\end{equation}
where $f_1, f_2$ and $f_3$ are some general functions, $\vec{x}(t)$ is the hidden state of the Expert at time $t$, and $\vec{o}(t)$ is the vector of observations in time $t$. The output projection function $f_3$ provides a compressed representation of the Expert's hidden state to its parents, which is then processed as their observations.  

We expect that there will be many Experts with highly overlapping (or nearly identical) receptive fields on each layer, which is motivated by the following two points:
\begin{itemize}
	\item Typically there will be multiple independent processes generating every localized part of the observation vector. So it might be beneficial to model them independently in multiple Experts.
    \item Since the Experts will learn in an unsupervised way, it is useful to have multiple alternative representations of the same observation $\vec{o}$ in multiple Experts. It might even be necessary in practice, since there is no one good representation for all purposes. Other Experts in higher layers can then either pick a lower-level Expert with the right representation for them or use outputs of multiple Experts below as a distributed representation of the problem (which has a higher capacity than a localized one~\cite{Hinton86}).
\end{itemize}

\subsection{Resulting Requirements on the Expert} \label{s:resultingRequirements}

As discussed in the previous section, the local model in each Expert might need to violate the Markov property and will never exactly correspond to a Generator sub-process. Thus, the goal of the Expert is not to model the input observations perfectly by itself, but to process them so that its output data is more informative about the environment than its inputs, and the Experts following in the hierarchy can make their own models more precise. 

In order to be able to successfully stack layers of multiple Experts on top of each other, the output of Expert $y$ has to use a suitable representation. This representation has to fulfill two seemingly contradictory requirements:

\begin{itemize}

\item It preserves spatial similarity of the input (see e.g. the Similar Input Similar Code (SISC) requirement in~\cite{Rinkus2014a} or Locality Sensitive Hashing (LSH)~\cite{Indyk}). In this case, the architecture should be able to hierarchically process the spatial inputs, even if there is no temporal structure that could be learned\footnote{Note that in the case where the output of the Spatial Pooler is a one-hot vector, the spatial similarity can be preserved only on the level of multiple experts, which together produce a locality-sensitive binary sparse code representing the input observation(s).}.

\item It should disambiguate two identical inputs based on their temporal (or top-down/lateral) context. The amount of context information added into the output should be weighted by the certainty about this context.

\end{itemize}

In the current implementation, we address this by converting the continuous observations into a discrete hidden state (based on the spatial similarity), which is then converted again into a (more informative) continuous representation on the output where the continuity captures the information obtained from the context inputs. It does so by working in four steps:
\begin{enumerate}

\item \textbf{Separation} (disentanglement) of observations produced by different causes (sub-generators).
\label{s:resultingRequirements:1}
The expert has to partition the input observations in a way that is suitable for further processing. Based on the assumption that values in each part of the observation space are a result of multiple sub-generators/causes (see \cref{envA:parallel}), the Expert should prefer to recognize part of the input generated by only one source. This can be achieved for example via spatial pattern recognition (parts of the observation space which correlate with each other are probably generated by the same source) or by using multiple Experts looking at the same data (see Appendix \ref{a:disentangled}). Alternative ways to obtain well disentangled representations of observations generated by independent sources are discussed in~\cite{Higgins, Thomas, Spratling2017}.

\item \textbf{Compression} (abstraction, associative learning).
\label{s:resultingRequirements:2}
Ideally, each expert should be able to parse high-dimensional continuous observations into discrete and localist (i.e. semi-symbolic) representations that are suitable for further processing. This can be done by associating parts of the observation together, which itself is a form of abstraction, and by omitting unimportant details. It is performed based on suitable criteria (e.g. associations of inputs from different sources seen frequently together) and under given resource constraints (e.g. a fixed number of discrete hidden states). This way, the expert efficiently partitions continuous observations into a set of discrete states based on their similarity in the input space.

\item \textbf{Expansion} (augmentation).
\label{s:resultingRequirements:3}
Since the input observations (and consequently the hidden states) can be ambiguous, each Expert should be able to augment information about the observed state so that the output of the Expert is less ambiguous and consists of Markov chains of lower order. This can be resolved e.g. by adding temporal, top-down  or lateral context~\cite{Laurent2016}.

\item \textbf{Encoding}.
\label{s:resultingRequirements:4}
The observed state augmented with the additional information has to be encoded. This encoding should be in a format which converts spatial and temporal similarity observed in the inputs, and similarity obtained from other sources (context), into a SISC/LSH space of the outputs.  thus enabling Experts higher in the hierarchy to do efficient separation and compression. 
\end{enumerate}

By iterating these four steps, a hierarchy of Experts is gradually converting a suboptimal or ambiguous model learned in the first layer of Experts into a model better corresponding to the true underlying HMM at a higher level. These mechanisms allow the architecture to partially compensate for the frequent inconsistencies between the hidden Generator and Learned Model topologies. Further improvements could potentially be based on distributed representations and forward/backward credit assignment (similar to~\cite{Rinkus2014a,Narendra2003,George2017a}).

\section{Description of the Prototype Architecture} \label{s:description}

At a high level, the passive architecture consists of a hierarchy of Experts (where $E_i^j$ denotes $i$-th expert in $j$-th layer), whose purpose is to match the hierarchical structure of the world (depicted in \cref{f:generator}) as closely as possible, as described in \cref{s:resultingRequirements}.

Separation (step \ref{s:resultingRequirements:1}) is solved on the level of multiple Experts looking at the same data and is described in more detail in Appendix \ref{a:disentangled}. Unless specifically stated, this version of disentanglement is not used in the experiments described in \cref{s:experiments}. 

Compression (step \ref{s:resultingRequirements:2}) is implemented by clustering input observations $\vec{o}(t)$ from a lower level (either another expert or some sensoric input) using the k-means\footnote{In the prototype, we cluster the data via k-means for simplicity and better interpretability, but there are no restrictions on how the compression is performed in general.}. The Euclidean distance from an input observation to known cluster centers is computed and the winning cluster is then regarded as the hidden state $\vec{x}(t)$ (see \cref{e:longerHmm}). This part is called the \textit{\qq{Spatial Pooler}}  (SP) (terminology borrowed from~\cite{Hawkins2006}).

The hidden state $\vec{x}(t)$ for the current time step is then passed to the next module called the \textit{\qq{Temporal Pooler}} (TP), which performs Expansion (step \ref{s:resultingRequirements:3}). It partitions the continuous stream of hidden states into sequences of (as in Layered Hidden Markov Models)\textemdash Markov chains of some small order $m>1$, and publishes identifiers of the current sequences and their probabilities.
It does so by maintaining a list of sequences and how often they occur. As it receives cluster representations\footnote{In the prototype, this is in the form of a 1-hot vector representing the index of the cluster.} from the SP, the TP learns to predict to which cluster the next input will belong. This prediction is calculated from how well the current history matches with the known sequences, the frequency that each sequence has occurred in the past, and any contextual information from other sources, such as neighboring Experts in the same layer, parent Experts in upper layers, or some external source from the environment.  

\begin{figure}[ht]
    \centering
    \includegraphics[width=1 \linewidth]{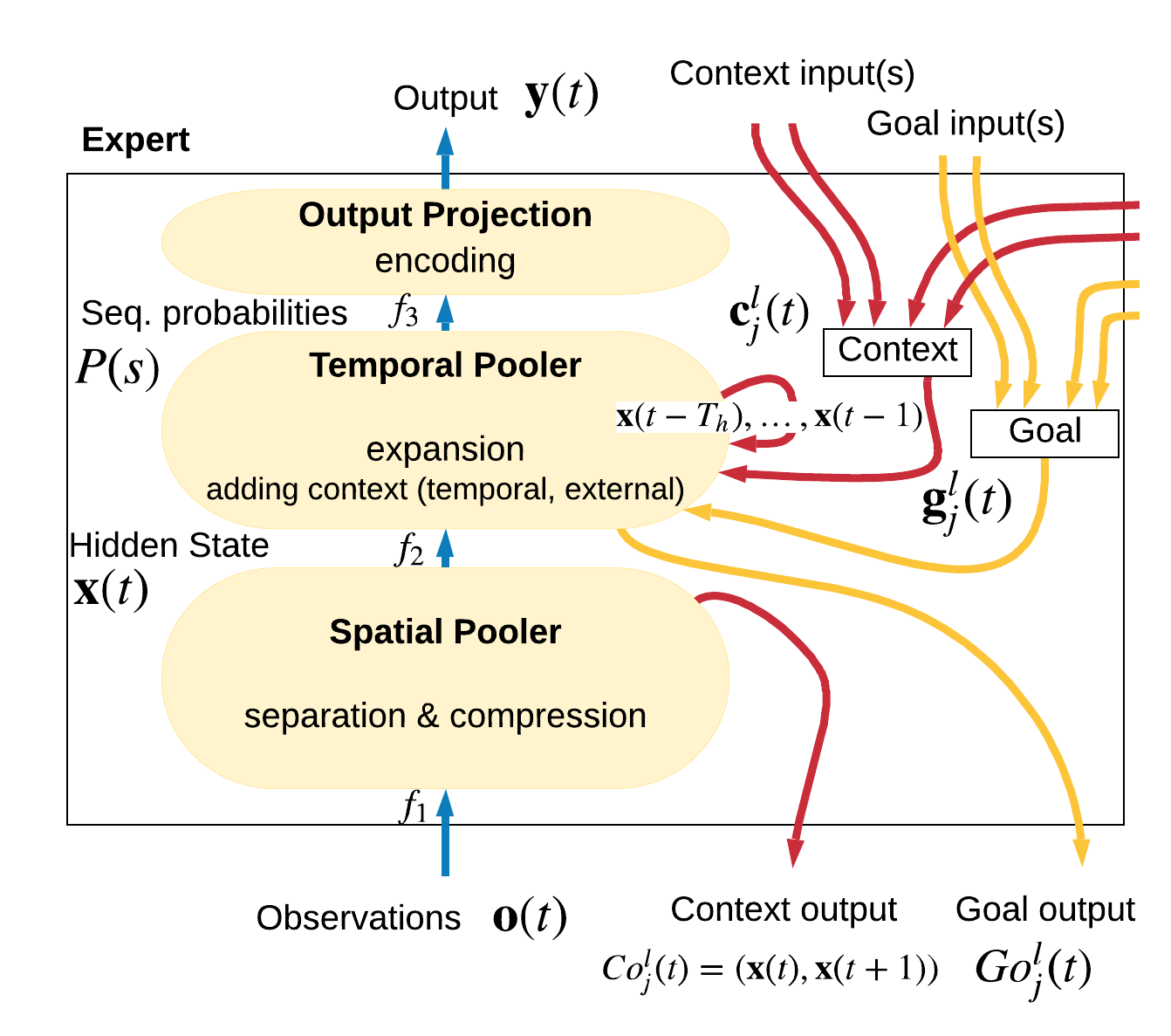}
    
    \caption{A High-level description of the Expert and its inputs/outputs. The observations are converted by the Spatial Pooler into a one-hot vector $\vec{x}(t)$ representing cluster center probabilities. The Temporal Pooler computes probabilities of known sequences $P(S)(t)$, which are then projected to the output. The external context $\vec{c}_j^l(t)$ is received from top-down and lateral connections from other Experts. The corresponding goal vector $\vec{g}_j^l(t)$ is used to define a high-level description of the goal state. The Context output of the Expert typically conveys the current cluster center probabilities, while the Goal output represents a (potentially actively chosen) preference for the next state expressed as expected rewards. This can be interpreted as a goal in the lower levels or used directly by the environment (see Appendix \ref{s:goalDirected}).}     
    \label{f:expert}
\end{figure}

Encoding (step \ref{s:resultingRequirements:4}) is implemented via \textit{Output Projection}. The idea is to enrich the winning cluster (what the Expert has observed) with temporal context (past and predicted events). This way, the Expert is able to decrease the order of the Markov chain of recognized states. It is done by calculating the probability distribution over which sequences the TP is currently in, and subsequently, by calculating a distribution over the predicted clusters for the next input. This prediction is combined with the current and past cluster representations to create a \textit{projection} $\vec{y}(t)$ over the probable past, present, and future states of the sequence. This projection is passed to the SP of the next Experts in the hierarchy. See \cref{f:expert} for a diagram illustrating the dataflow.

The TP runs only if the winning cluster in the SP changes which results in an event-driven architecture. The SP serves as a quantization of the input so that if the input does not change enough, the information will not be propagated further.

The context is a one-way communication channel between the TP of an Expert and the TP(s) of the Expert(s) below it in the hierarchy. This context serves two purposes: First, as another source of information for a TP when determining in which sequence it is. And second, as a way for parent Experts to communicate their goals to their children\footnote{As the parents are not connected directly to the actuators, they have to express their desired high-level (abstract) actions as goals to their children which then incorporate these goals into their own goals and propagate them lower. Experts on the lowest levels of the hierarchy are connected directly to actuators and can influence the environment.}, depicted in \cref{f:expert} and explained in Appendix \ref{s:goalDirected}.

The context consists of three parts: 1) the output of the SP (i.e. the cluster representation), 2) the next cluster probabilities from the TP, and 3) the expected value of any rewards that the architecture will receive if in the next step, the input falls into a particular cluster (interpreted as goals)\footnote{In \cref{f:expert}, the goals are shown as separate from the context for clarity.}. 

In order to influence the environment, the Expert first needs to choose an action to perform, which is the role of the active architecture. The goal is a vector of rewards that the parent expects the architecture will receive if the child can produce a projection $\vec{y}(t+1)$ which will cause the parent SP to produce a hidden state $\vec{x}(t+1)$ corresponding to the index of the goal value.

An expert receiving a goal context computes the likelihood of the parent getting to each hidden state using its knowledge of where it presently is, which sequences will bring about the desired change in the parent, and how much it can influence its observation in the next step $\vec{o}(t+1)$. It rescales the promised rewards using these factors, and adds knowledge about its own rewards it can reach. Then it calculates which of its hidden states lead to these combined rewards. From here, it publishes its own goal $Go$ (next step maximizing the expected reward), and if it interacts directly with the environment picks an action to follow\footnote{The action of bottom level Experts at $t-1$ is provided on $\vec{o}(t)$ from the environment, so the picking of an action is equivalent to taking the cluster center of the desired state and sampling the actions from the remembered observation(s). See Appendix \ref{s:actions} for more details.}.

A much more detailed description of the architecture, its mechanics, and principles can be found in the Appendix.

\section{Experiments} \label{s:experiments}

This relatively simple architecture combines a number of mechanisms. The general principles of the ToyArchitecture has broad applicability to many domains. This can be seen in the variety of experiments which can be devised for it, from making use of local receptive fields for each Expert, to processing short natural language statements. 

Rather than going though all of them, this section will instead show some selected experiments which focus on demonstrating and validating the functionality of the mechanisms described in this paper. The experiments were performed in either BrainSimulator\footnote{\url{https://www.goodai.com/brain-simulator}} or TorchSim\footnote{\url{https://github.com/GoodAI/torchsim}}. The source code of the ToyArchitecture implementation in TorchSim is freely available alongside TorchSim itself.

\subsection{Video Compression - Passive Model without Context}

We demonstrate the performance of a single Expert by replicating an experiment from~\cite{Laukien2017}. The input to the Expert is the video from the paper with a resolution of $192 \times 192 \times 3$, composed of 434 frames. 

The experiment demonstrates a basic setting, where the architecture just passively observes and the data has a linear structure with local dependencies. Therefore, a single Expert is able to learn a model of this data with only the passive model and without needing context, as detailed in Appendix \ref{s:passive}.

The Expert has 60 cluster centers and was allowed to learn 3000 sequences of length 3, where the lookbehind (how far in the past the TP looks to calculate in which sequence it is currently in) is $T_b=2$ and the lookahead (how many future steps the TP predicts) is $T_f=1$. Both the SP and TP were learning in an on-line fashion. The video is played for 3000 simulation steps during training (through the video almost 9 times). The cluster centers are initialized to random small values with zero mean.

\begin{figure}[ht]
    \centering
    \includegraphics[width=\linewidth]{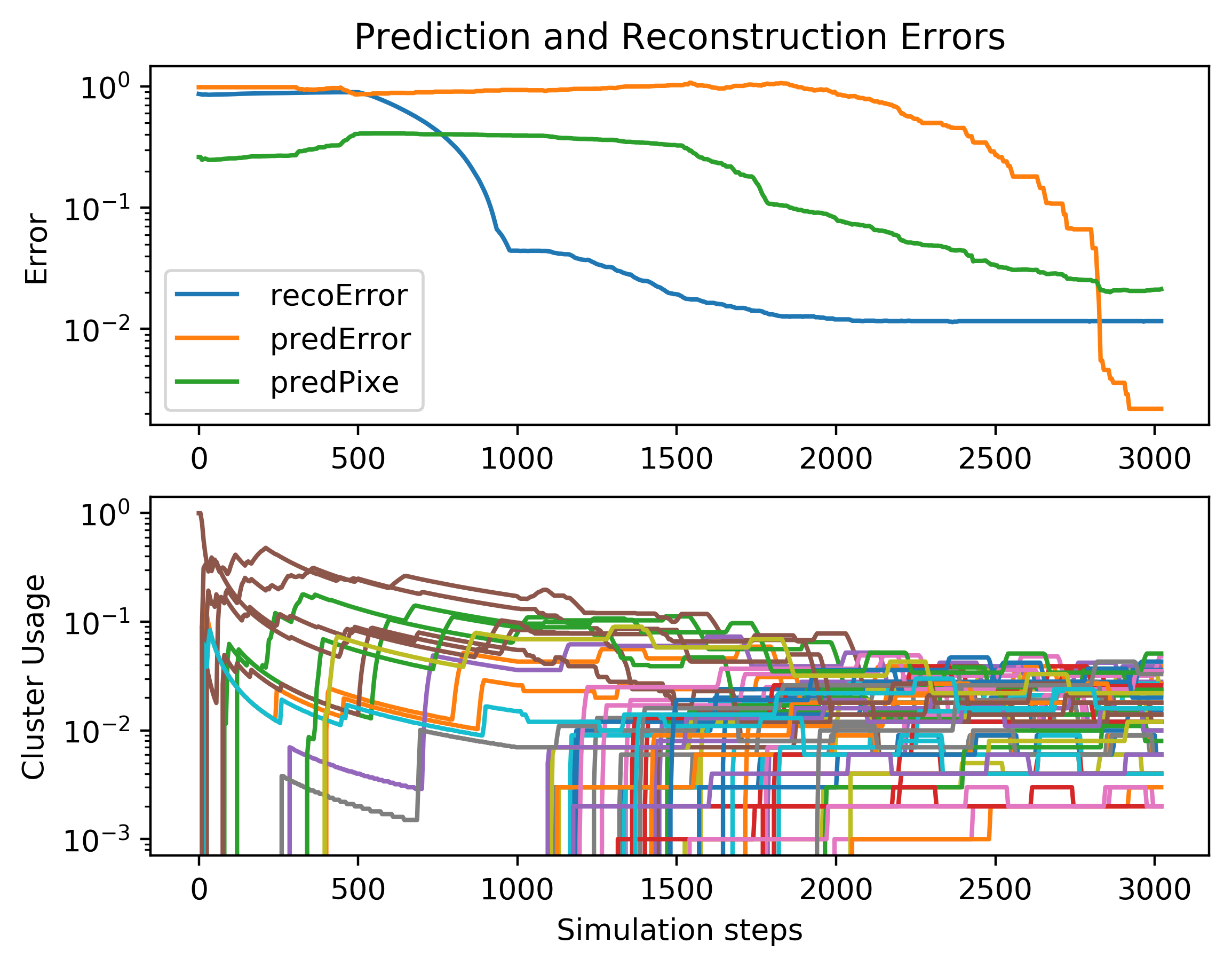}
    \caption{\textbf{Top graph:} reconstruction and prediction errors during the course of on-line training of the Expert on the video. \textbf{Bottom graph}: cluster usage (moving window averaged), where each line represents the percentage of time each cluster is active. 
    Both parts of the Expert (the Spatial Pooler and Temporal Pooler) learn on-line (internal training batches are sampled from the recent history). The reconstruction error (in the observation/pixel space: \textit{'recoError'}) decreases first, because the Spatial Pooler learns to cluster the video frames. This causes an overall decrease of prediction error in the observation/pixel space: \textit{'predPixe'}. Note that around step 1700, the prediction error in the observation space decreases, despite the fact that the internal prediction error increases.  This is because the changes in the SP representation degrade the sequences learned by the TP. Around step 2000, the learned spatial representation (clustering) is stable (cluster usage shows that all clusters have data) and therefore the inner representation of temporal dependencies starts to improve. Around step 3000, the Temporal Pooler predicts perfectly. The \textit{'predError'} is measured as a prediction error in the hidden space (on the clusters).}
    \label{f:bird:errors_log}
\end{figure}

\cref{f:bird:errors_log} shows the course of: reconstruction error, prediction error in the pixel space, and prediction error in the hidden representation\footnote{In all cases the error is computed as the sum of squared differences between the reconstruction (prediction) and the ground truth.}.

First, the SP learns cluster centers to produce $\vec{x}(t)$ given a video frame at time $t$. In the beginning, only a small number of cluster centers are trained, therefore the winning cluster changes very sporadically (all of the data is chunked into just a small number of clusters). Since the TP runs only if the winning cluster of the SP changes, this results in a situation where the data for the TP changes very infrequently, which means that the TP learns very slowly. This can be seen around step 1000 in \cref{f:bird:errors_log}, where the reconstruction error converges between $10^{-2}$ and $10^{-1}$. At this point, boosting\footnote{A mechanism which moves unused cluster centers towards the populated ones with the largest variation in their data points, see Appendix \ref{s:passive}.} starts to have an effect, which results in all clusters being used.

\begin{figure}[ht]
    \centering
     	\subfigure[Sim. step 1000]{
        	\includegraphics[width=0.27\linewidth]{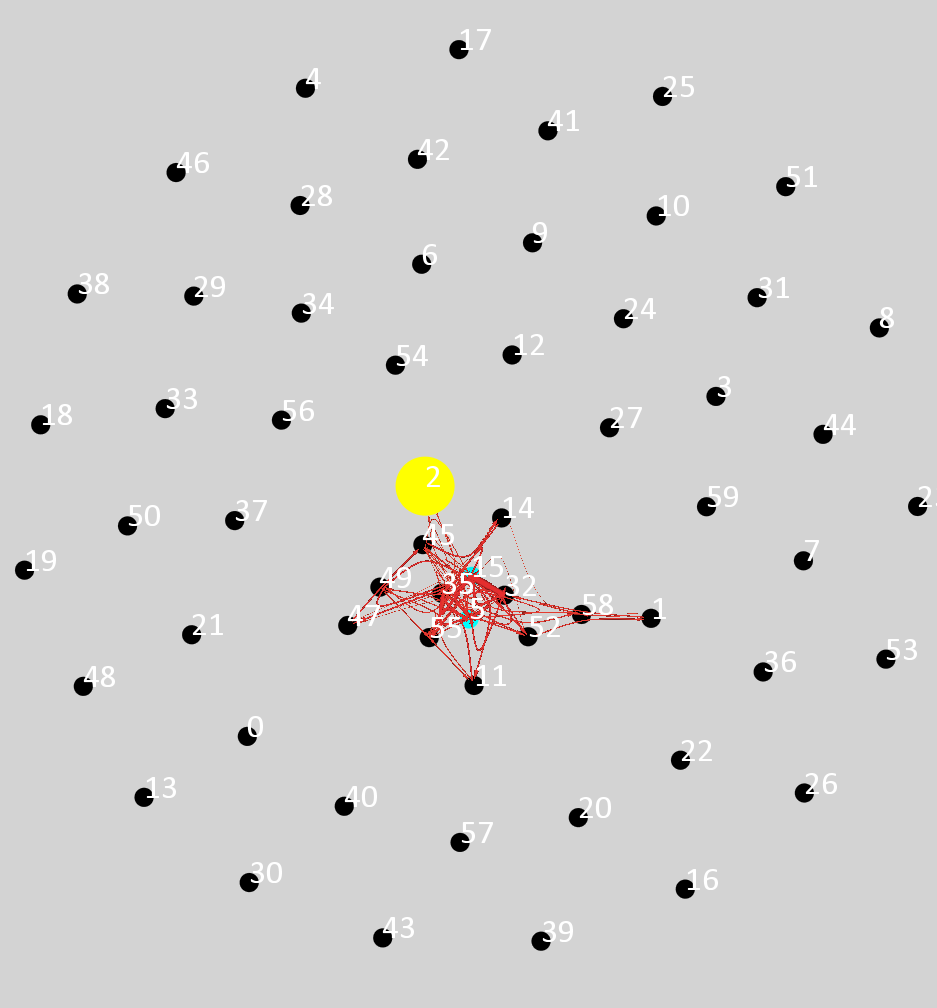}
    	}
     	\subfigure[Sim. step 1500]{
        	\includegraphics[width=0.27 \linewidth]{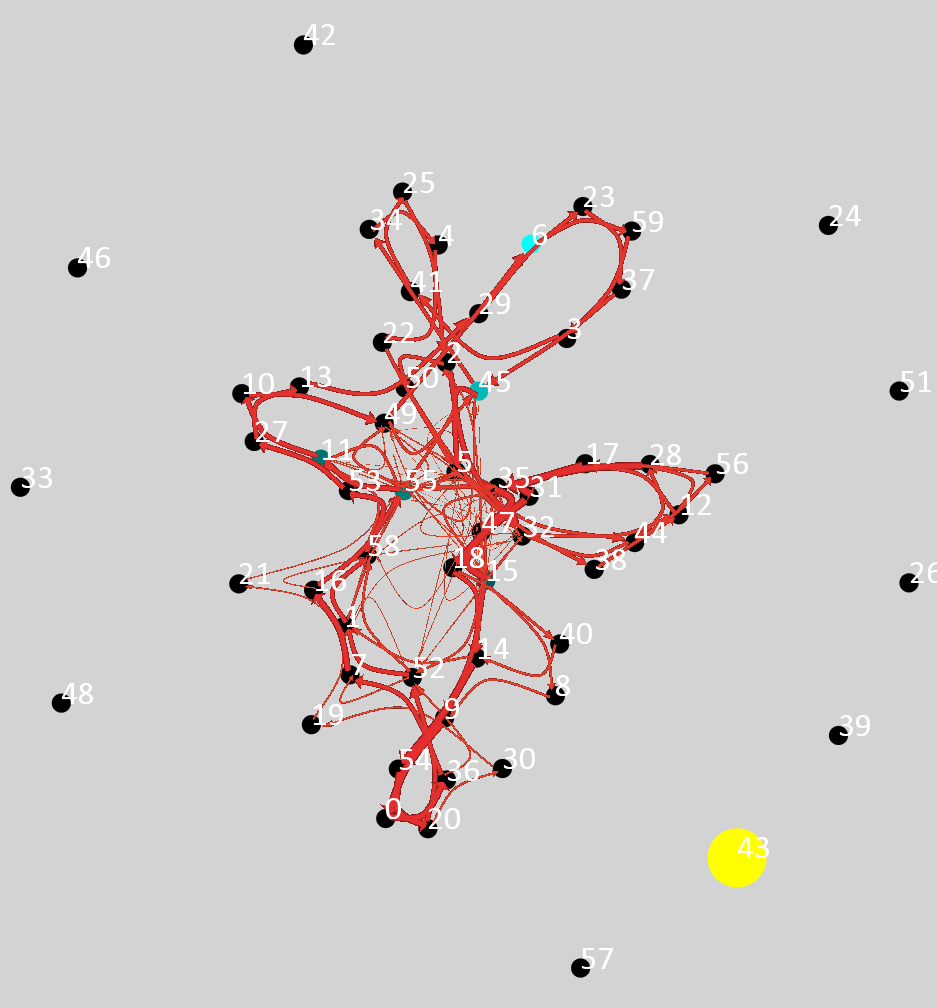}
    	}
        \subfigure[Sim. step 3000]{
        	\includegraphics[width=0.27 \linewidth]{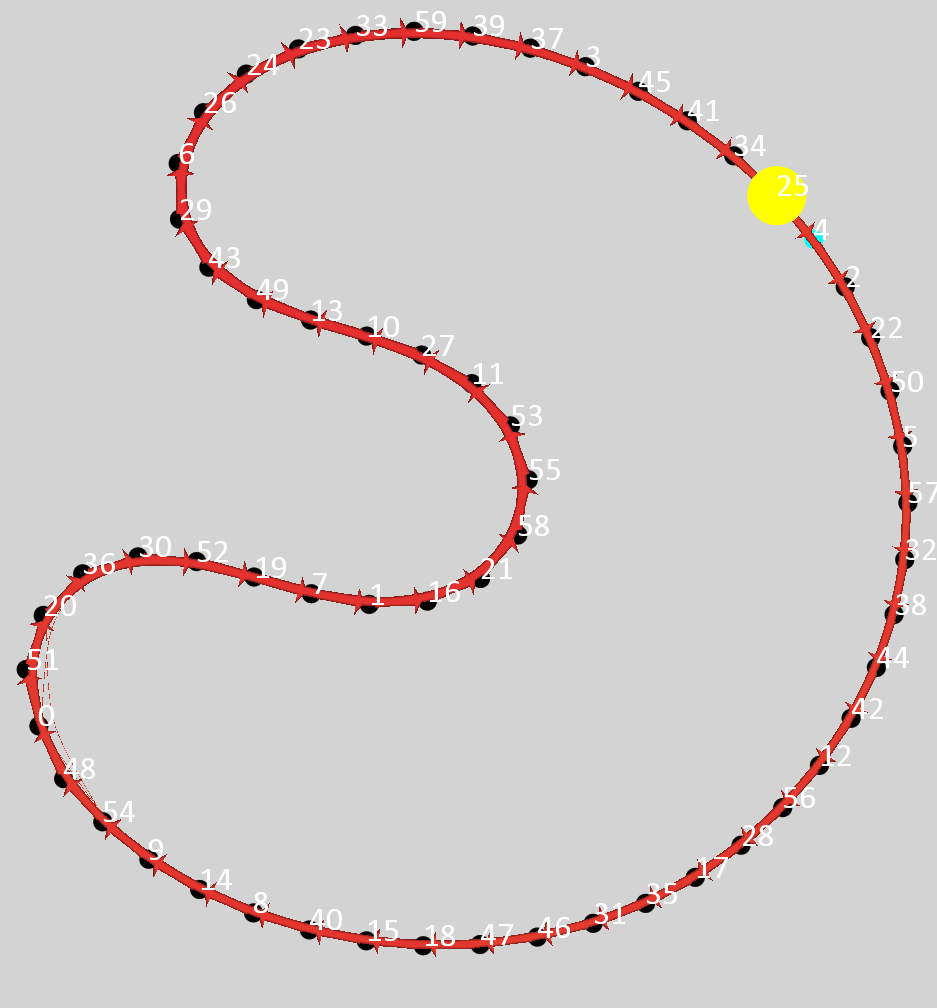}
    	}
    	
        \caption{Convergence of the Temporal Pooler's transition probabilities. Each point is a cluster center. The Expert learns sequences of 3 consecutive cluster centers. At the beginning of the simulation, only several cluster centers are used, the Temporal Pooler learns transitions between currently used clusters. Finally, when all the cluster centers are used for some time, the Temporal Pooler converges and learns the linear structure of the video.}
    
    \label{f:bird:tp-sequence}
\end{figure}

The larger the number of clusters in use, the more often the TP sees an event ($\vec{x}(t)$ changes), and the more frequently it learns. In the last stage of the experiment, the prediction errors start to converge towards zero.  

\begin{figure}[ht]
    \centering
    \includegraphics[width=\linewidth]{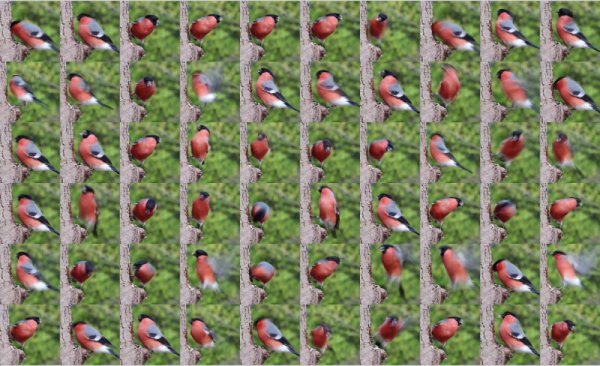}
    \caption{Resulting cluster centers after learning. Each of the 60 clusters corresponds to approximately 7 frames in the video. The cluster $x(t)$ is active when the nearest 7 frames in the video are encountered. This results in spatial (but also temporal) compression. Note that one Expert is not designed to learn such a large input. Rather, multiple Experts with local receptive fields should typically process the input collaboratively.}      
    \label{f:bird:sp}
\end{figure}

This results in a trained Expert, which can recognize the current observation (compute $\vec{x}(t)$) and reconstruct it back. The learned spatial representation is shown in \cref{f:bird:sp} and the convergence of the temporal structure is shown in \cref{f:bird:tp-sequence}. 

As a result, given two\footnote{Just the current state would be enough in this experiment, since the learned temporal structure has the Markov property.} consecutive hidden states $\vec{x}(t-1:t)$, the Expert can predict the next hidden state $\vec{x}(t+1)$ and reconstruct it in the input space. This process can be seen in a supplementary video\footnote{Video generated by the Expert: \url{http://bit.ly/2um5zyc}}. The first part of the video shows how the Expert can recognize and reconstruct current observation and predict the next observation. The second part of the video (48 seconds in) shows the case where the prediction of the next frame is fed back to the input of the Expert. This shows that the Expert can be `prompted' to play the video from memory. The spatio-temporal compression caused by the clustering and event-driven nature of the TP results in a faster replay of the video as only significant changes in the position of the bird are clustered differently and thus remembered as events by the TP.

\textbf{Discussion:} This experiment demonstrates the capability of on-line learning on linear data of one Expert using the passive model without context. The Expert first learns to chunk/compress the input data into discrete pieces (modelling the hidden space) and then to predict the next state in this hidden space. The prediction can be then fed back to the input of the Expert, which results in replaying meaningful input sequences (where time is distorted by the event-driven nature of the algorithm). 

Two important remarks can be made here.
\begin{enumerate}
	\item The reconstruction error converges to small values fast, but only a small fraction of clusters is used at that moment. After this first stage, all the clusters start to be used. This change improves the reconstruction error slightly and allows the Temporal Pooler to start learning. This is relevant to~\cite{Schwartz-Ziv2017}, where it is argued that the internal structure of the network changes, even if it might not be apparent from the output.
	\item The prediction error in the pixel space decreases before the prediction error in the hidden space starts to decrease. The reason for this is that even if the Temporal Pooler predicts a uniform distribution over the hidden states $x(t+1)$ (i.e. the TP is not trained yet), all the cluster centers are moving closer towards the real data and thus the average prediction improves no matter what cluster is predicted.
\end{enumerate}

This experiment shows the performance of an Expert on video, but the same algorithm should process other modalities as well without any changes. It shows a trivial case, where the hidden space just has a linear structure (Markov process). The following experiment extends this to non-Markovian case, where the use of context is beneficial.

\subsection{Audio Compression -- Passive Model with Context}

\begin{figure}[ht]
    \centering
    \includegraphics[width=0.9\linewidth]{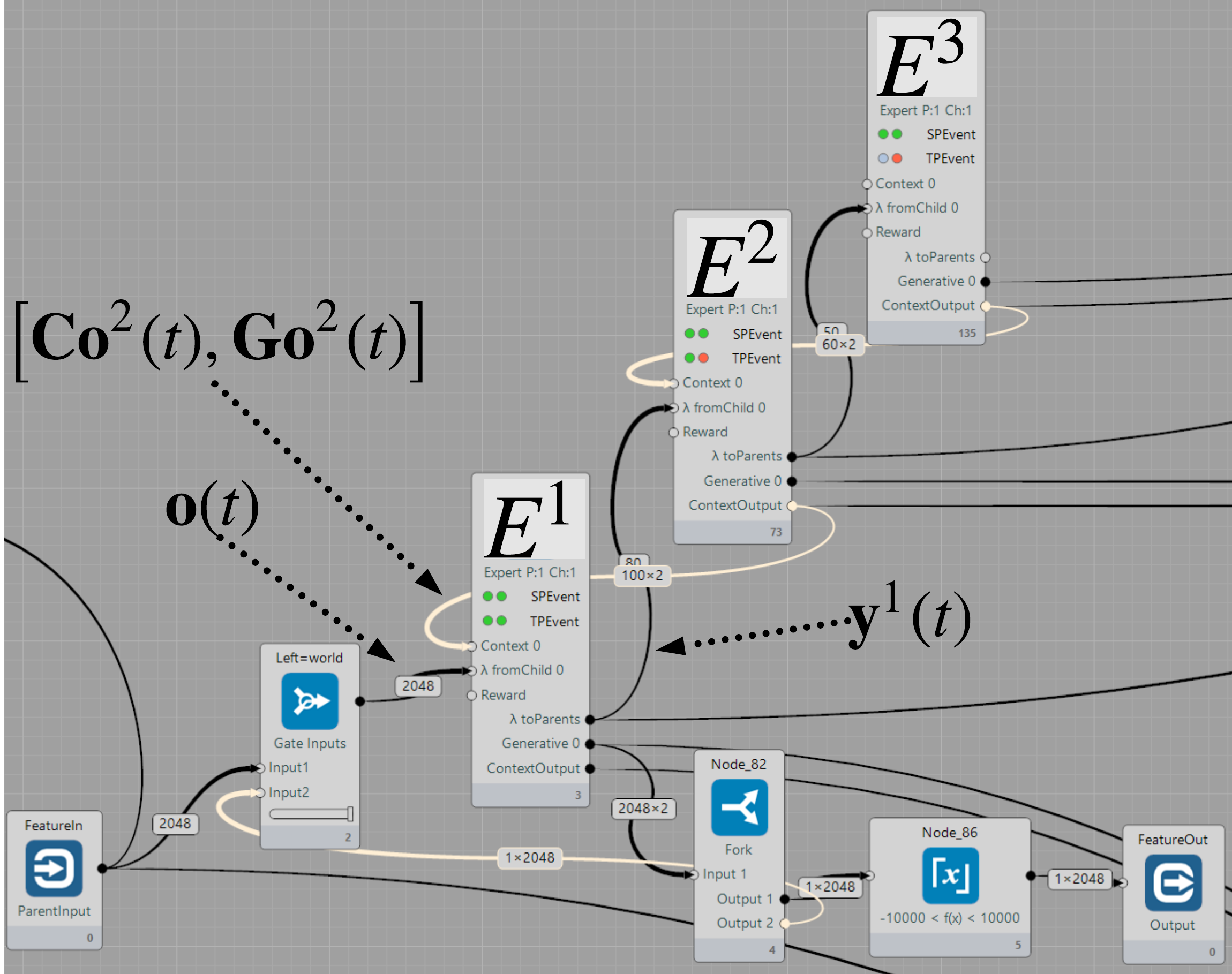}
    \caption{Setup of the experiment with context. $E^1$ receives the feature vectors on the $\vec{o}(t)$ input and receives the context vector $\large[\vec{Co}^2(t), \vec{Go}^2(t) \large]$ from its parent $E^2$, which helps it to resolve uncertainty in the Temporal Pooler. The same holds for higher level(s).}
    \label{f:audio:brain}
\end{figure}

This experiment demonstrates a simple layered interaction of three Experts connected in three layers, as depicted in \cref{f:audio:brain}. Its purpose is to demonstrate that top-down context provided by parent Experts helps improve the prediction accuracy at the lower levels.

The setup is the following: Expert $E^1$ in layer 1 processes\footnote{Normally, $E_i^j$ denotes the $i$-th Expert in the $j$-th layer. Since this experiment uses just one Expert per layer, the subscript $i$ will be omitted for clarity.} the observations $\vec{o}(t)$ and computes outputs $\vec{y}^1(t)$, the parent Expert $E^2$ in layer 2 processes the output vector of $E^1$: $\vec{y}^1(t)$ as its own observation and produces the context vector. This context vector is used by the $E^1$ to improve the learning of its own Temporal Pooler as described in Appendix \ref{s:externalContext}. The same is done for the third layer.

The input data to the architecture is an audio file with a sequence of spoken numbers 1-9. The speech is converted by Discrete Fourier Transform into the frequency domain with 2048 samples. Each time step, one sample with 2048 values is passed as an observation $\vec{o}(t)$ to $E^1$. The original audio file is available on-line\footnote{Original audio file with labels: \url{http://bit.ly/2HxdTUA}}.

All the Experts in the hierarchy share the same number of available sequences (2000), and the lookahead $T_f=1$. The Expert $E^1$ which processes raw observations has 80 cluster centers and a sequence length $m=3$. Its parent Expert $E^2$ has 50 cluster centers with a sequence length $m=5$. The most abstract Expert $E^3$ has 30 clusters and can learn sequences of length $m=5$.

\begin{figure}[ht]
    \centering
    \includegraphics[width= \linewidth]{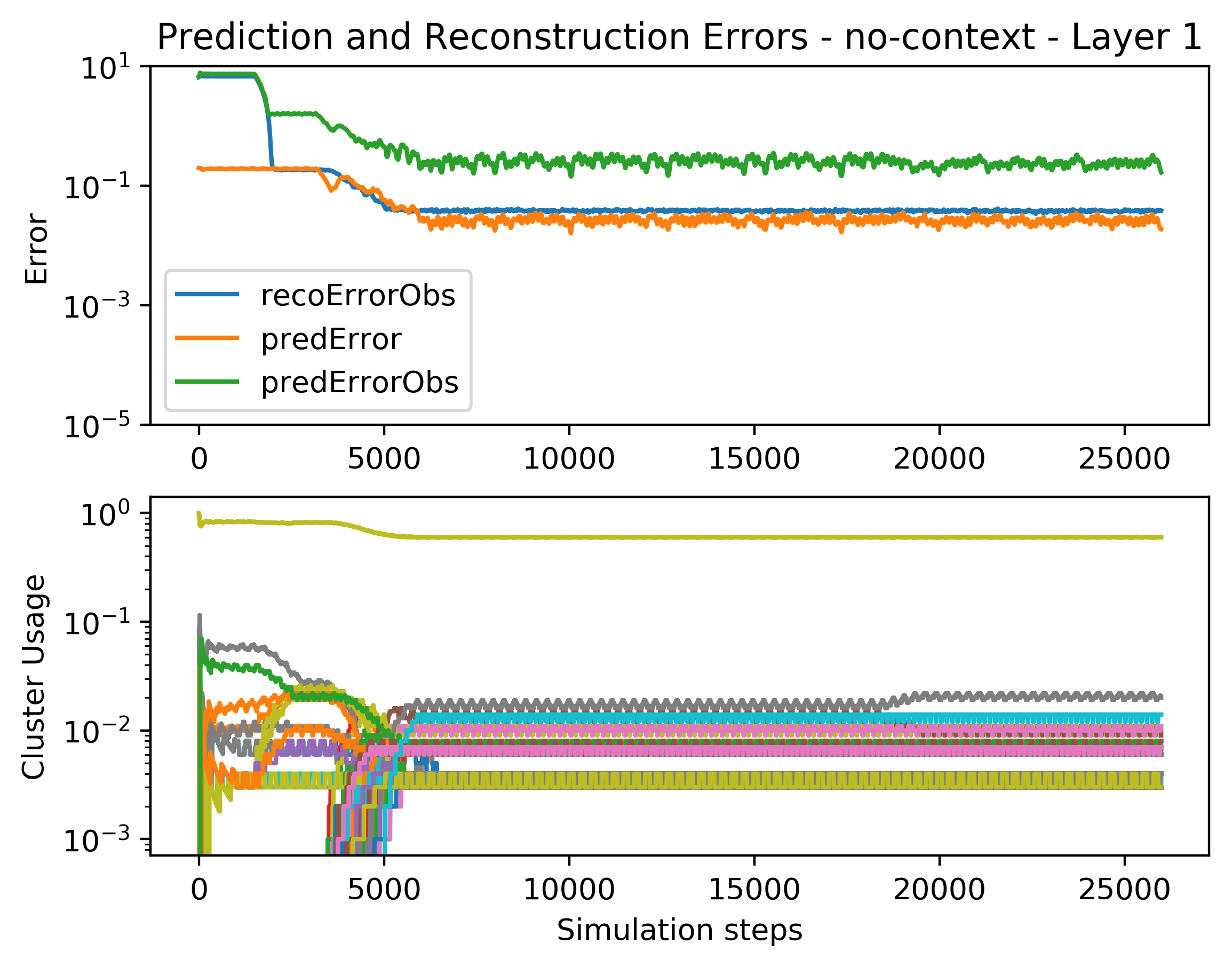}
    \caption{Convergence of the Spatial Pooler's reconstruction error (\textit{recoErrorObs}) and Temporal Pooler's prediction error both in the observation (\textit{predErrorObs}) and hidden (\textit{predError}) space. The graph below shows cluster usage in time: one of the clusters is used much more often, probably representing a silent part. The prediction error converges to a relatively high value, since the Expert is unable to learn the model correctly by itself.}
    \label{f:a:noContext}
\end{figure}

\begin{figure}[ht]
    \centering
     	\subfigure[Transitions in $E_1^1$]{
        	\includegraphics[width=0.295\linewidth]{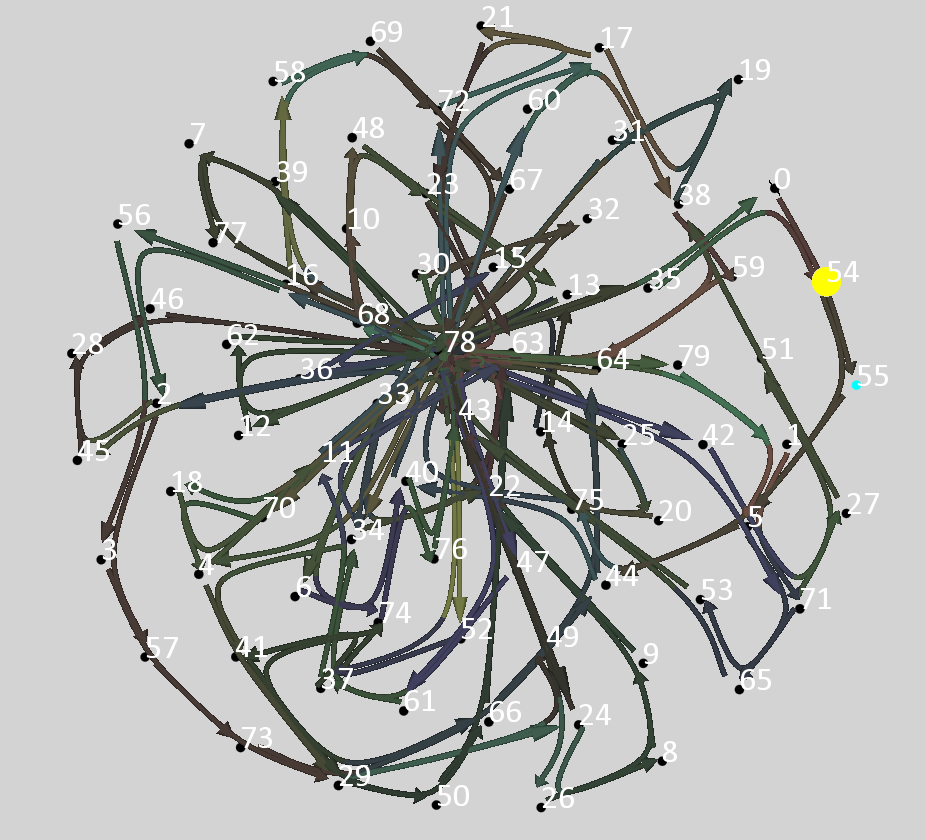}
        	\label{f:a:transitions:1}
    	}
     	\subfigure[Transitions in $E_1^2$]{
        	\includegraphics[width=0.295 \linewidth]{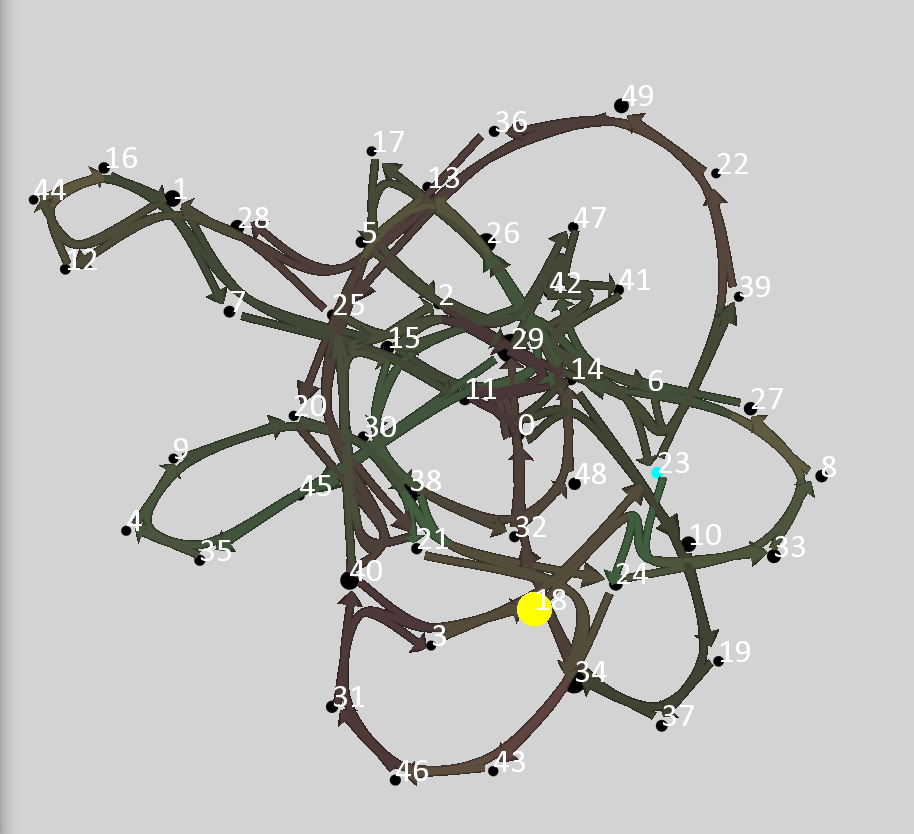}
    	}
        \subfigure[Transitions in $E_1^3$]{
        	\includegraphics[width=0.295 \linewidth]{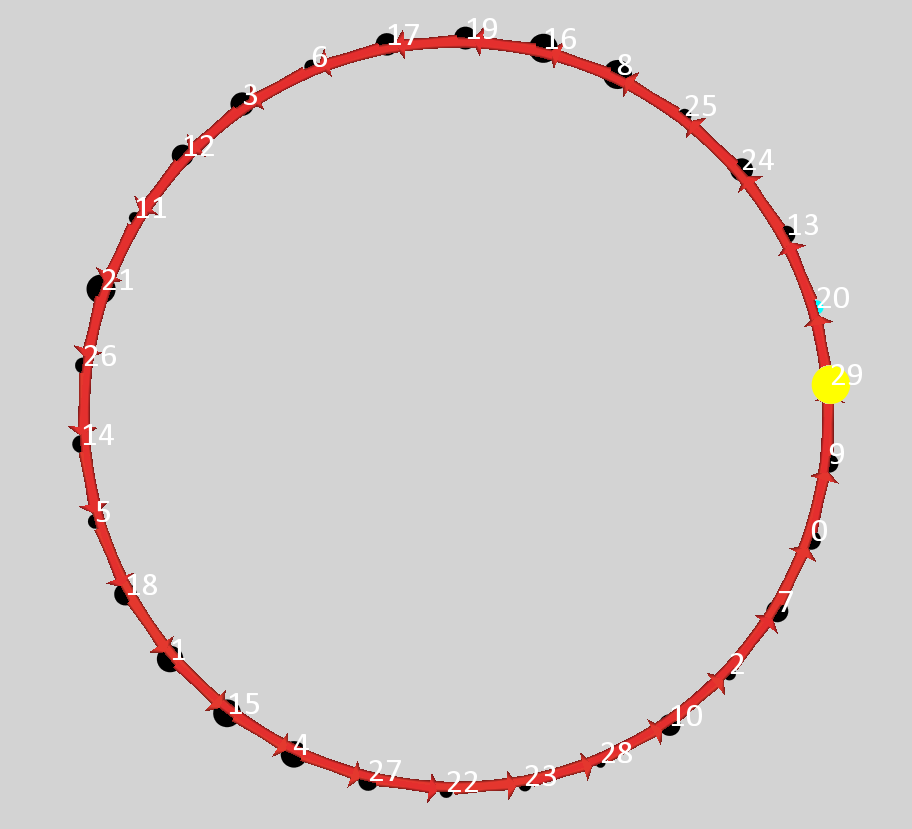}
    	}
        \caption{Resulting learned sequences in all the Experts. It can be seen how the output projections to $y^i(t)$ help to adaptively compress predictable parts of the input. The higher in the hierarchy, the lower the order of the Markov chain the Experts process. On the top of the hierarchy, the order is 1 and for the Expert $E_1^2$ the sequence of hidden states has a linear structure.} 
    \label{f:a:transitions}
\end{figure}

The results of a baseline experiment with just the bottom Expert $E^1$ are shown in \cref{f:a:noContext}. After training both Spatial and Temporal Poolers, it can be seen that the sequences of hidden states $\vec{x}^1(t)$ are highly non-Markovian (\cref{f:a:transitions:1}). The order of the Markov chains is higher than the supported maximum of $m-1=2$. After connecting the prediction\footnote{In this simulation, the GreedyWTA function was applied on the prediction.} to the Expert's input as a new observation $\vec{o}(t)$, the Expert is almost able to reconstruct two words, but is stuck in a loop of these two\footnote{The audio generated by one Expert without context available is located at: \url{http://bit.ly/2W7OXpO}, after some time the prediction starts failing.}. The reason of is that many sequences are going through several clusters which correspond to relative silence. In these states, the Expert does not have enough temporal context to determine in which direction to continue.

But if we connect several Experts in multiple layers above each other, the parent Experts provide temporal context $Co^l(t)$ to the Experts below. Since the Experts higher in the hierarchy represent the process as a Markov chain of lower order (see \cref{f:a:transitions}), the context vector provided by them serves as extra information according to which the low-level Expert(s) can learn to predict correctly. Due to the event-driven nature of each Expert, the hierarchy naturally starts to learn from the low level towards the higher ones. Once learned, the average prediction error on the bottom of the 3-layer hierarchy is lower compared to the baseline 1-Expert setup. 

After connecting the bottom Expert in a closed loop, like in the previous experiment, the entire hierarchy is able to replay the audio correctly. The resulting audio can be found on-line\footnote{Audio generated by a hierarchy of 3 Experts: \url{http://bit.ly/2FrnFWg}.} and the representation is shown in \cref{f:a:transitions}. 

\begin{figure}[ht]
    \centering
    \includegraphics[width= \linewidth]{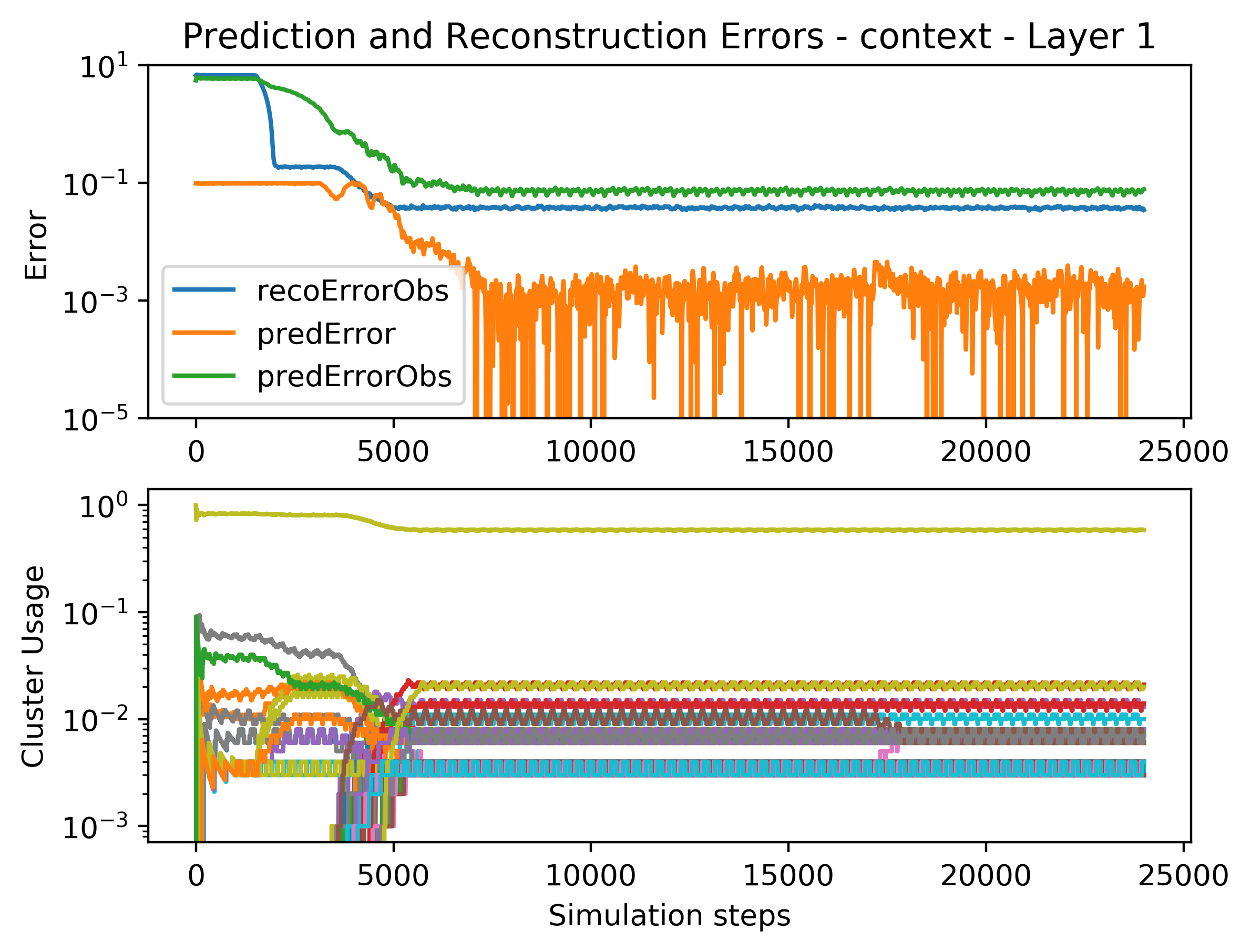}
    \caption{Error and cluster usage charts tracking the convergence of the bottom Expert $E^1$ in the presence of the context signal. Compare with the baseline in \cref{f:a:noContext} which does not use the context. See \cref{f:a:noContext} for a description of the plotted lines. Since the processing of the SP is not influenced by the context, the SP works identically as in the baseline case (e.g. the cluster usage and SP outputs are the same in both experiments).}
    \label{f:a:context1}
\end{figure}

\begin{figure}[ht]
    \centering
    \includegraphics[width= \linewidth]{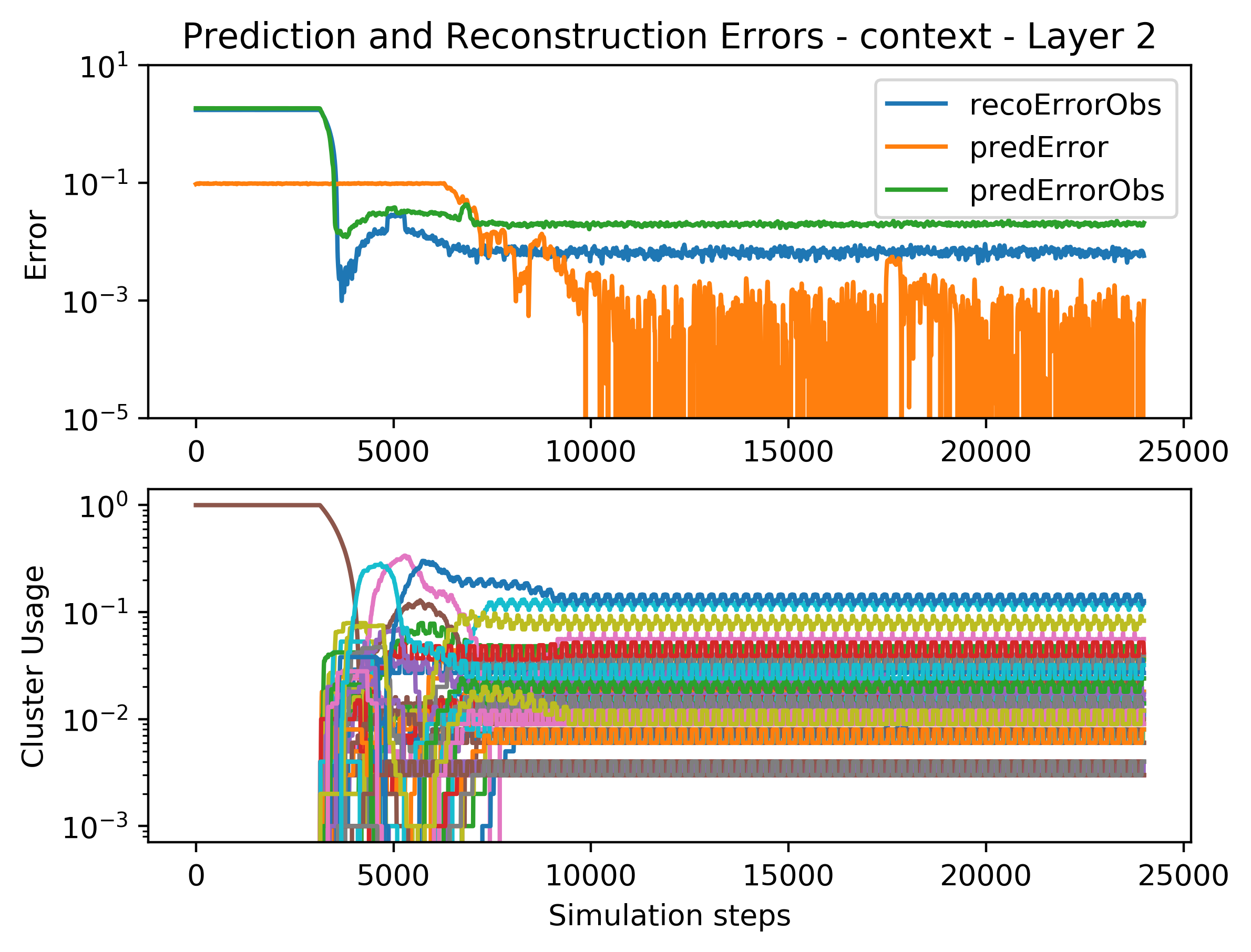}
    \caption{Error and cluster usage charts tracking the convergence of $E^2$. See \cref{f:a:noContext} for a description of the plotted lines.} 
    \label{f:a:context2}
\end{figure}

\begin{figure}[ht]
    \centering
    \includegraphics[width= \linewidth]{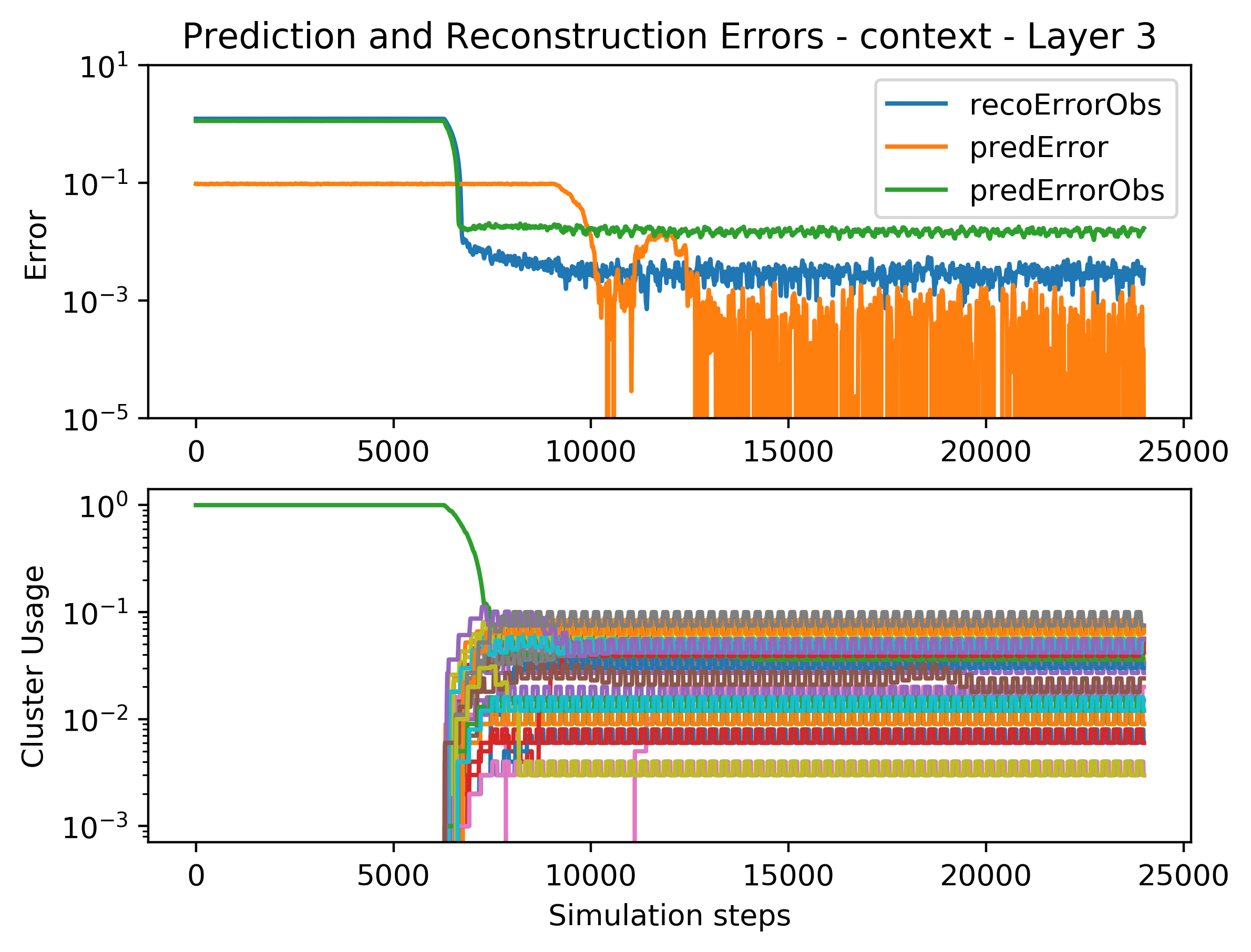}
    \caption{Error and cluster usage charts tracking the convergence of $E^3$. See \cref{f:a:noContext} for a description of the plotted lines.} 
    \label{f:a:context3}
\end{figure}

Figures \ref{f:a:context1}, \ref{f:a:context2}, and \ref{f:a:context3} show the convergence of the Spatial and Temporal Poolers for each Expert in the hierarchical setting.  The Spatial Pooler in the bottom layer behaves identically as in \cref{f:a:noContext}, but here, the Temporal Pooler can use the top-down context to decrease its prediction errors significantly. The cluster usage graphs show the effect of increasingly abstract representations. In layers 2 and 3, there is no explicit cluster for silence as in the first layer, because those silences cannot be used to predict the next number, and so are disregarded.

Note that the event-driven processing in the Experts, the architecture implements  adaptive compression in the spatial and temporal domain on all levels. This is exhibited as either speeding up the video in the preceding experiment, or speeding up the resulting generated audio in this experiment.

\textbf{Discussion:} The experiment has shown how the context can be used to extendthe ability of a single Expert to learn longer term dependencies. It has also shown that the hierarchy works as expected: higher layers form representations that are more compressed and have lower orders of Markov chains. The activity on higher layers can provide useful top-down context to lower layers, and these lower layers can leverage it to decrease their own prediction error.

\subsection{Learning Disentangled Representations}

This experiment illustrates the ability of the architecture to learn disentangled representations of the input space. In other words, this is the ability to recover hidden independent generative factors of the observations in an unsupervised way. Such an ability may be vital for learning grounded symbolic representations of the environment~\cite{Higgins, Bengio2017}. In the prototype implementation, the ability to disentangle the generative factors is implemented via a predictive-coding-inspired mechanism (described in Appendix \ref{a:disentangled}), and is limited only to the input being created by an additive combination of the factors.

The experiment shows how a group of two Experts can automatically decompose the visual input into independently generated parts of the input. And to naturally learn about each of them separately, without any domain-specific modifications.

\begin{figure}[ht]
    \centering
    \includegraphics[width=\linewidth]{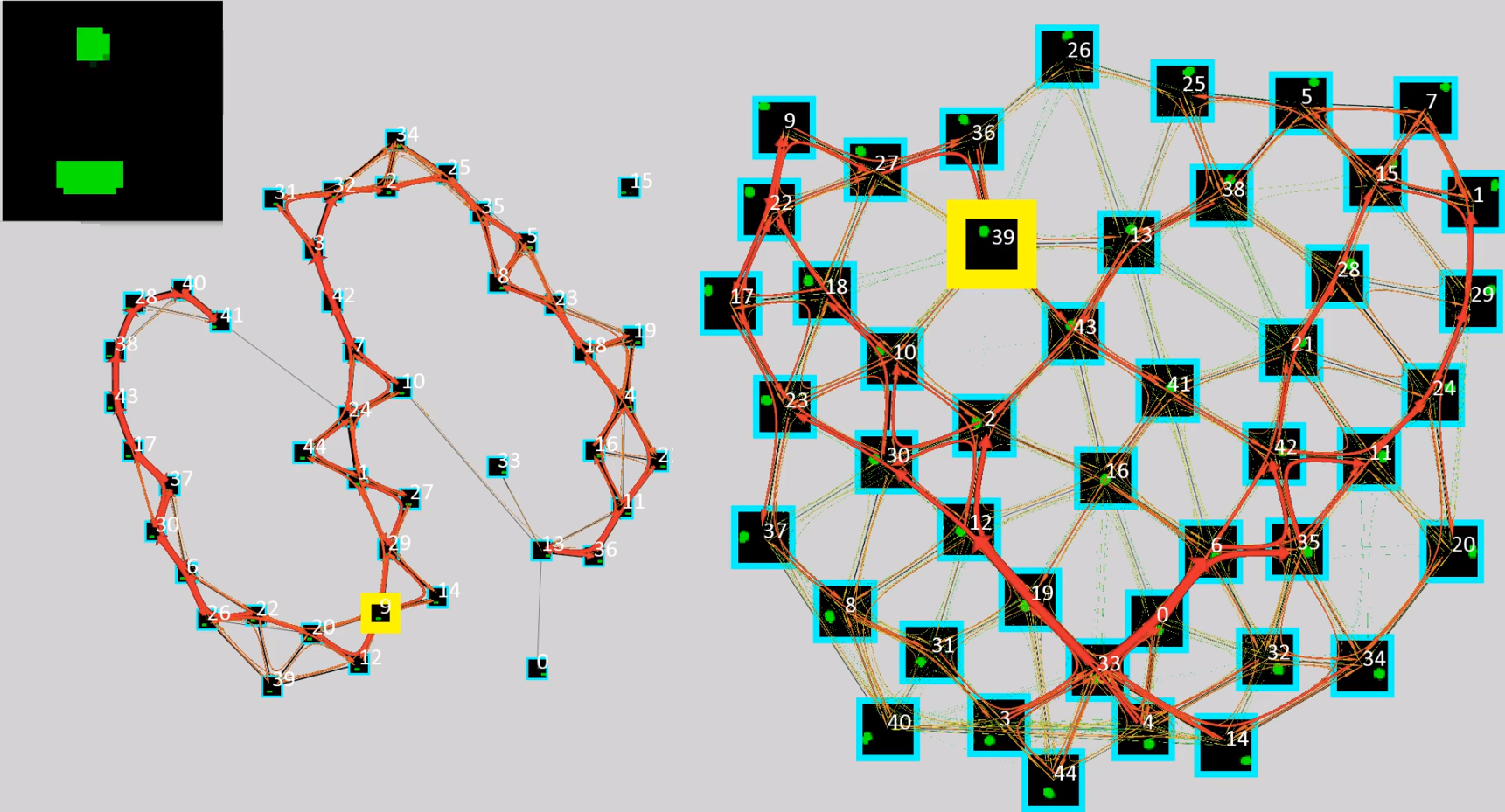}
    \caption{\textbf{Top left:} the current visual input (pong, with ball and paddle). The image shows the representation of two objects learned in an unsupervised way by two Experts competing for the same visual input. The learning automatically decomposes the observations into two independent parts. The independent parts in this case correspond to the paddle (left) and the ball (right). By representing each object in a separate Expert, each is able to learn the simple temporal structures governing the behavior of its object independently of the other, leaving the learning of structures resulting from the interaction of the objects to higher and more abstract layers. From the representation it can be easily seen that the paddle moves just in one axis (linear structure discovered by the TP), while the ball moves through the entire 2D space (grid). The current position of the ball and the paddle are shown in yellow, each cluster center is overlaid with the visual input it represents.}
    \label{f:a:pong}
\end{figure}

The input is a sequence of observations of a simple gray-scale version of the game pong (shown in the top left in  \cref{f:a:pong}). The ball moves on realistic trajectories and the paddle is moved by an external mechanism so that it collides with the ball around $90\%$ of the time. 

The experiment shows how a simple competition of two Experts for the visual input can lead to the unsupervised decomposition of observations into independent parts. Here, there are two mostly independent parts on the input, therefore the Spatial Pooler of one Expert represents one part (paddle), the other Expert the other part (ball). The resulting representations are shown in \cref{f:a:pong}. The rest of the architecture works without any modification, therefore each of the Temporal Poolers learn the behavior of just a single object\footnote{See the illustrative video of the inference at \url{http://bit.ly/2CvXnQv}}. Representing states of each of the objects independently is much more efficient than representing each state of the scene at once.

\textbf{Discussion:} Although this simple mechanism is not as powerful as DL-based approaches~\cite{Higgins}, it is interpretable and considerably simpler. It was experimentally tested that such a configuration is able to disentangle up to roughly 6 independent sources of input. In case the number of latent generative factors of the environment is smaller than the number of competing Experts $M<N$, then the group of Experts forms a sparse distributed representation of the input. It is a topic for further research if application of this simple mechanism on each layer of the hierarchy\footnote{As with each mechanism in the ToyArchitecture, we expect the workload to be distributed among all Experts, closely interacting with other mechanisms, and performed using simple algorithms rather than being localized in one part of the architecture and solving the problem all at once.} could overcome its limitations and achieve results comparable to deep neural networks.

\subsection{Simple Demo of Actions}

The purpose of this experiment is to show the interplay of most of the mechanisms in the architecture. A small hierarchy of two Experts has to learn the correct representation of the environment on two levels of abstraction, then use this representation to explore, discover a source of reward, learn its ability to influence the environment through actions and then collect the rewards. The passive model works identically to the previous experiments, and addition the active parts of the model are enabled. Moreover, all the active parts of the model should be backwards compatible, which means that this configuration of the network should work also on the previous experiments, even though there are no actions available.

\begin{figure}[ht]
    \centering
    \includegraphics[width=0.6 \linewidth]{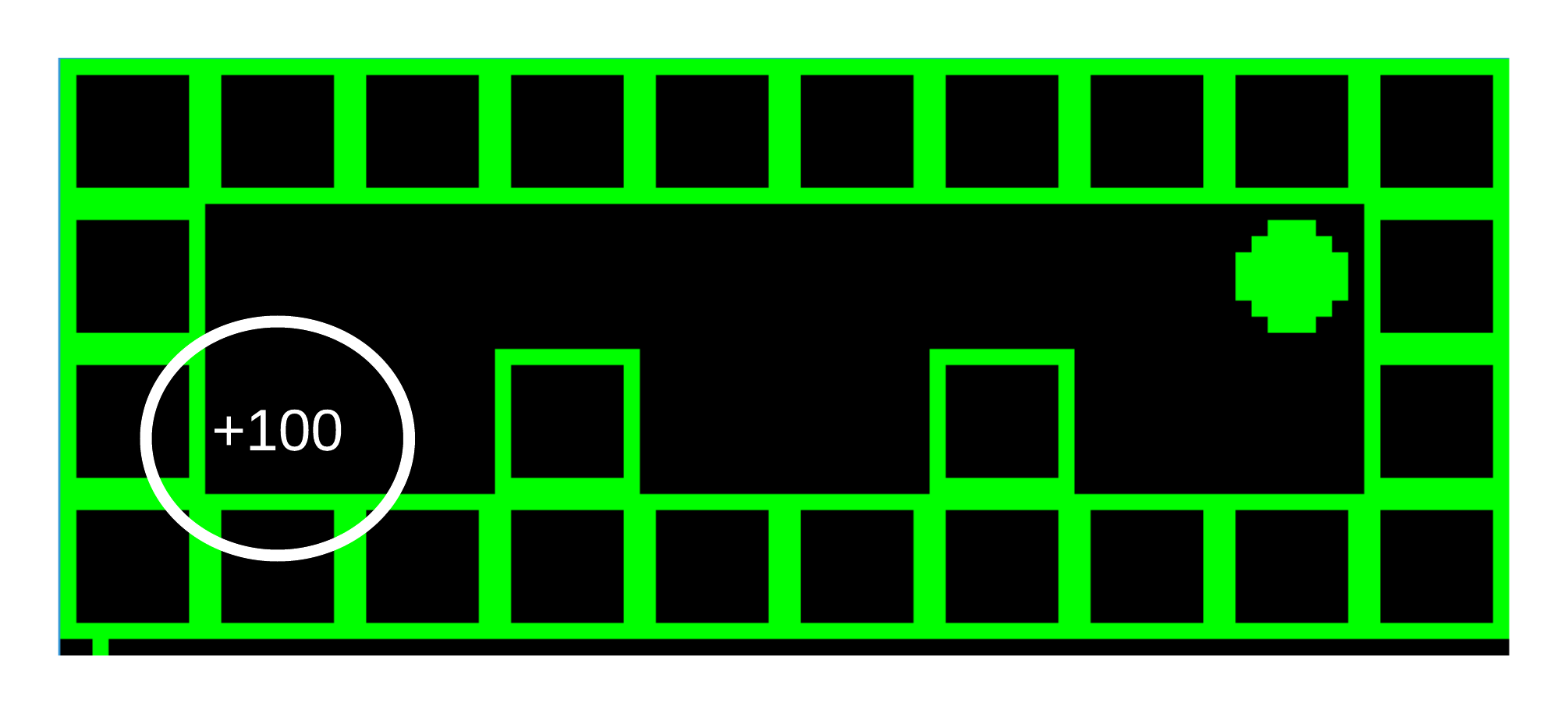}
    \caption{The initial state of the world, with the agent represented as the green circle and the reward tile highlighted by the authors.}
    \label{f:simple-GD}
\end{figure}

This experiment uses a hierarchy of two Experts to find and continuously exploit reward in a simple gridworld. Each time the agent obtains the reward, its position is reset to a random position where there is no reward. The reward location is fixed, but visually indicated. The agent must therefore explore tiles to find it and remember the position. \cref{f:simple-GD} pictures the initial state of the world.

The agent itself consists of two Experts connected in a narrow hierarchy similar to the one depicted in \cref{f:audio:brain}. Expert $E^1$ has 44 cluster centers, a sequence length of 5 and lookahead of 3, and $E^2$ has 5 clusters with 7 and 5 for sequence length and lookahead respectively. As stated in appendices \ref{s:actions} and \ref{s:goalDirected} the agent sees the action on the input (the one pixel tall 1-hot vector in the bottom left of  \cref{f:simple-GD}), and all levels receive reward (100, in this case) when the agent steps onto the reward tile. 

With a lookahead of 3, $E^1$ can `see' the reward only 2 actions into the future (the reward is given when the agent is reset, so it is effectively delayed by 1 step). Expert $E^2$ meanwhile clusters sequences from $E^1$, so that it has a longer `horizon' over which to see. Expert $E^2$ therefore has to guide $E^1$ to the vicinity of the reward tile by means of the context and goal vectors.

\begin{figure}[ht]
    \centering
    \includegraphics[width=1.0 \linewidth]{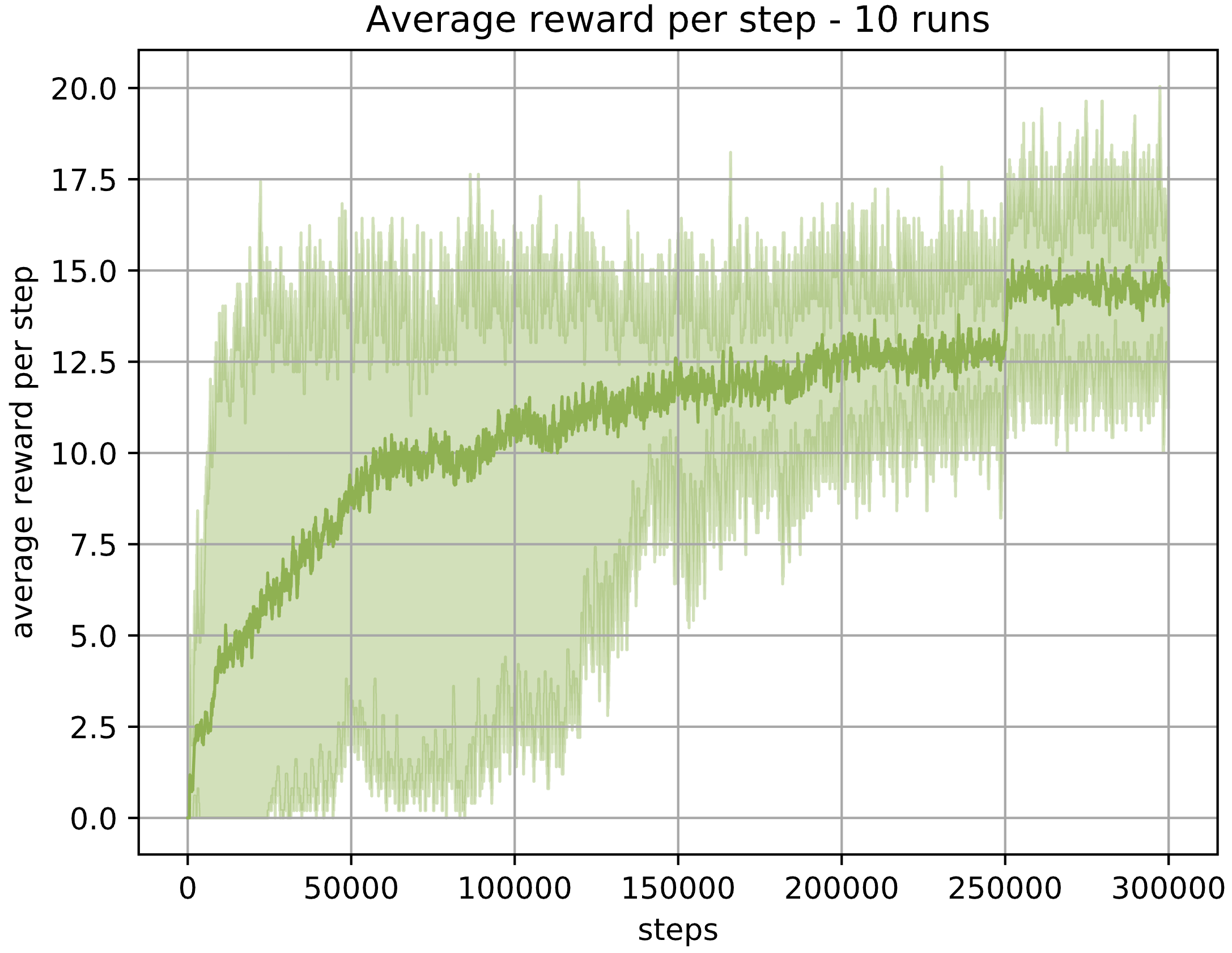}
    \caption{The average (minimum and maximum) collected reward per step across 10 runs. Learning and exploration was disabled after the step 250,000.}
    \label{f:gd_chart}
\end{figure}

The results of 10 independent runs measured by average reward per step is presented in \cref{f:gd_chart}. As one would expect the average reward increases as time goes on, indicating that the agent has learned where the reward is, and is actively following its learned path to that reward.

\begin{figure}[ht]
    \centering
    \includegraphics[width=0.3 \linewidth]{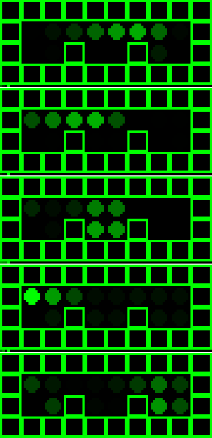}
    \caption{An interpretation of the clusters of $E^2$, projected through the clusters of $E^1$ and into the input space. Expert $E^2$ clusters spatial and temporal information from $E^1$, so its clusters represent a superposition of states of $E^1$.}
    \label{f:GD-simple-hirearchical}
\end{figure}

A particularly good example of $E^2$ clustering is in \cref{f:GD-simple-hirearchical}. This shows that $E^2$ had created clusters where temporally contiguous projections from $E^1$ are spatially clustered together. So that if we were to overlap these 5 images there would be a contiguous `line' of agent positions from anywhere in the environment to right beside the reward tile. 
\textbf{Discussion:} This experiment demonstrates that the hierarchical exploration and goal-directed mechanisms are functional and, when trained appropriately, allow an Expert hierarchy to find rewards and follow goals. However, when the clustering is done poorly (as has been the case for at least one run of the experiment), the model encounters a lot of difficulty. Since the model is constantly learning, the cluster centers might find a global (local) optima or continuously drift in time. Therefore, incentivising a `good' clustering without domain specific knowledge is currently an open question and will be mentioned further in \cref{s:conclusion}.   

\subsection{Actions in Continuous Environments}

The current design of the architecture supports not only discrete environments, but was also tested in continuous environments with continuous actions. The last experiment serves as a simple illustration of this and is similar to another experiment of the authors of \cite{Laukien2017}\footnote{Link to the video of the original experiment: \url{https://bit.ly/2XVZmXF}.} 

\begin{figure}[ht]
    \centering
    \includegraphics[width=0.5 \linewidth]{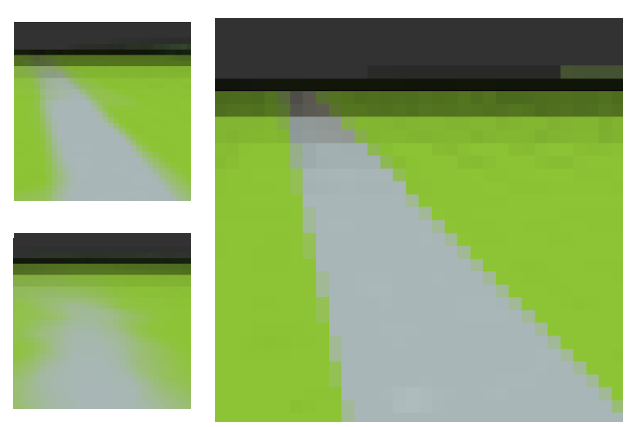}
    \caption{An example of first-person visual input to the Expert. \textbf{Right:} current visual input; \textbf{top left:} reconstruction of the current cluster (the part which corresponds to the visual input); \textbf{bottom left:} reconstruction of selected next cluster center (the part which corresponds to the visual input, the other part is taken as an action to be executed). The Expert is predicting that it will turn left in the next step, and therefore the track will correspondingly be more in the center of the visual field.}
    \label{f:a:driver-input}
\end{figure}

\begin{figure}[ht]
    \centering
    \includegraphics[width=1 \linewidth]{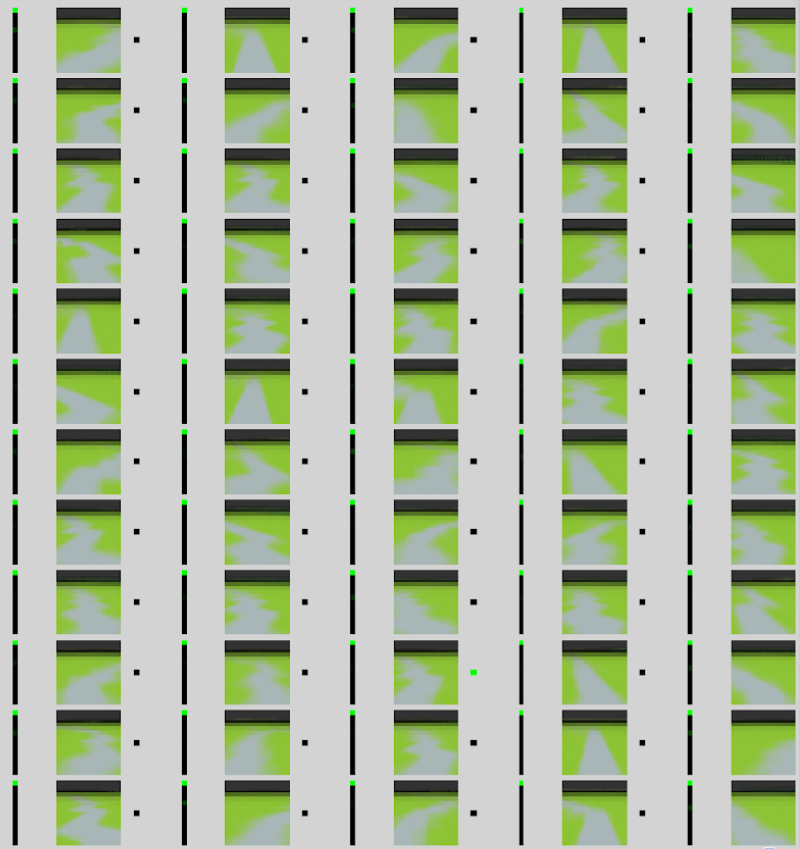}
    \caption{An example of cluster centers learned in $E^1$ after puppet-learning (showing ground-truth action). The task can be solved pretty well by a reactive agent (stimulus $\rightarrow$ response policy). As a consequence of this, each cluster center represents some visual input and its corresponding learned action. Training in a RL setting, where the reward is given for staying on the road, leads to very similar cluster centers.}
    \label{f:a:driver}
\end{figure}

The environment is a simple first-person view of a race track. The goal is to stay on the road and therefore to drive as fast as possible.

The topology is composed of just one Expert $E^1$ which receives a visual image and a continuous action (the top bit is forward, and then there are barely visible slight turning actions below) stacked together.

\textbf{Discussion:} The single Expert was able to learn to drive on a road in a so called puppet-learning setting, where the correct (optimal) actions are shown (a human drove through the track manually several times). But it was also able to learn correct behavior in a RL setting, where just the visual input and a reward signal (for staying on the road) was provided. Despite the fact that the learned representation is simple and seems to be on the edge of memorization, the agent was able to generalize well and was able to navigate also on previously unseen tracks (with the same colors). An example of agent autonomously navigating in the racing track is online\footnote{Autonomous navigation of the agent on the race track: \url{http://bit.ly/2OgkVO5}.}.

These five experiments suggest that hierarchical extraction of spatial and temporal patterns is a relatively domain-independent inductive bias that can create useful models of the world in an unsupervised manner, forming a basis for sample efficient supervised learning. The same basic architecture has been tested on a variety of tasks, exhibiting non-trivial behaviour without requiring domain specific information, nor huge volumes of data on which to train. 
\section{Discussion and Conclusions}\label{s:conclusion}

This paper has suggested a path for the development of general-purpose learning algorithms through their interpretability. First, several assumptions about the environments were made, then based on these assumptions a decentralized architecture was proposed and a prototype was implemented and tested. This architecture attempts to solve many problems using several simple and interpretable mechanisms working in conjunction. The focus was not on performance on a particular task, it was rather on the generality and the potential to provide a platform for sustainable further development.

We presented one relatively simple and homogeneous system which is able to model and interact with the environment. It does this using the following mechanisms:
\begin{itemize}
    \item extraction of spatio-temporal patterns in an unsupervised way,
    \item formation of increasingly more abstract and more informative representations,
    \item improvement of predictions on the lower levels by means of the context provided by these abstract representations,
    \item learning of simple disentangled representations,
    \item production of actions and exploration of the environment in a decentralized fashion,
    \item and hierarchical, decentralized goal-directed decision making in general.  
\end{itemize}

\subsection{Similar architectures}

There are many architectures/algorithms which share some aspects with the work presented here. The similarities can be found in the focus on unsupervised learning, hierarchical representations, recurrence in all layers, and the distributed nature of inference.

The original inspiration for this work was the PhD Thesis \qq{How the Brain Might Work}~\cite{D1987}. The hierarchical processing with feedback loops in ToyArchitecture is similar to \textit{CortexNet}~\cite{Canziani2017}, a class of networks inspired by the human cortex. There are also a lot of architectures that are more or less inspired by predictive coding \cite{Bastos2012,Spratling2017}, but they are focused on passively learning from the data. 

Many of these architectures are implemented in ANNs, using the most common neuron model. They are often similar in their hierarchical nature, such as the Predictive Vision Model \cite{Richert2016a}; a hierarchy of auto-encoders predicting the next input from the current input and top-down/lateral context. More recently, the Neurally-Inspired Hierarchical Prediction Network \cite{Qiu2019} uses convolution and LSTMs connected in a predictive coding setting. Several publications try to gate the LSTM cells in a manner inspired by cortical micro-circuits \cite{PonteCostaa}.

There are more networks that are loosely inspired by these concepts. The main idea is usually in the ability to have some objective in all layers, enabling the network to produce intermediate gradients which improves convergence and robustness. Examples of these are Ladder Networks \cite{Rasmus}, or the Depth-gated LSTM \cite{Yao2015}.

There are also networks that use their own custom model of neurons. These include the Hierarchical Temporal Memory (HTM) \cite{Hawkins2015}, the Feynman Machine \cite{Laukien} or Sparsey \cite{Rinkus2014a}.

A model inspired by similar principles was also able to solve CAPTCHA. It is the Recursive Cortical Network (RCN) \cite{George2017a}. It works on visual inputs that are manually factorised into shape and texture. Compared to other architectures mentioned here, it is based on probabilistic inference and therefore is closer to the hypothesis that the brain implements Bayesian inference \cite{Lee}.

There are fewer architectures that are also focused on learning actions. An example of a system implemented using deep learning techniques is Predictive Coding-based Deep Dynamic Neural Network for Visuomotor Learning \cite{Hwang2017}. It learns to associate visual and proprioceptive spatio-temporal patterns, and is then able to repeat the motoric pattern given the visual input. The Feynman Machine was also shown to learn and later execute policies taught via demonstration \cite{Laukien2017}. Despite the fact that both of the architectures are able to learn and execute sequences of actions, none of them currently support autonomous active learning. In contrast to the ToyArchitecture, the mechanisms for exploration and learning from rewards are missing. An architecture emphasizing the role of actions and active learning in shaping the representations is~\cite{Hay2018}. Similarly to the ToyArchitecture, actions are part of the concept representation and not just the output of the architecture.

A more loosely bio-inspired architecture is World Models \cite{Ha2018}. These combine VAE for spatial compression of the visual scene, RNNs for modeling the transitions between the world states, and a part which learns policies. Compared to the ToyArchitecture, this structure is only has a single layer (just one latent representation) and learns its policies using an evolutionary-based approach. Here, the interesting aspect is that after learning the model of the environment, the architecture does not need the original environment to improve itself. It instead `dreams' new environments on which to refine its policies.

Another deep learning approach focused on a universal agent in a partially observable environment is the MERLIN architecture~\cite{Wayne2018}. Based on predictive modelling, it tries to learn how to store and retrieve representations in an unsupervised manner, which are then used in RL tasks. Unlike the ToyArchitecure, it is a flat system where the memory is stored in one place instead of in a distributed manner.  

\subsection{Limitations and Future Work}

Despite promising initial results, the theory is far from complete and there are many challenges ahead. The performance of the model is partially sacrificed for interpretability, and in the current (purely unsupervised or semi-supervised setup) it is far behind its DL-based counterparts. It seems that the current biggest practical limitation of the model is that the Experts do not have efficient mechanisms to make the representation in other Experts more suitable for their own purposes (i.e. a mechanism which implements credit assignment through multiple layers of the hierarchy). There are some potentially promising ways how to improve this (either based on an alternative basis \cite{Rinkus2014a}, a DL-framework \cite{Qiu2019} or a probabilistic one \cite{George2017a}).

Another way to scale up the architecture would be to use multiple Experts with small, overlapping receptive fields (as discussed in \cref{s:hierarchicalPartitioning}), ideally in combination with a mechanism efficiently distributing the representations among them (see Appendix \ref{a:disentangled}). Our preliminary results (not presented in this paper) show that such redundant representations can not only increase the capacity of the architecture~\cite{Hinton86}, but also provide a population for evolutionary based algorithms of credit assignment. 

During development, empirical evidence suggested that a better form of lateral coordination (lateral context between Experts) is missing in the model, especially in the case of wide hierarchies with multiple experts on each layer processing information from local receptive fields. Examples of this can be seen in \cite{Laurent2016} and \cite{George2017a}.

Some mechanisms to obtain a grounded symbolic representation of the environment were tested in the form of disentanglement. It is not clear now whether these mechanisms would be scalable all the way towards very abstract conceptual representations of the world, or if there is something missing in the current design which would support abstract reasoning.

One of the big challenges in designing complex adaptive systems is in life-long or gradual learning; i.e. the ability to accumulate new non-i.i.d. knowledge in an increasingly efficient way\cite{Rosa2016}. The system has to be able to integrate new knowledge into the current knowledge-base, while not disrupting it too much. It should also be able to use the current knowledge-base to improve the efficiency of gathering new experiences. So despite that some of these topics are partially covered by the architecture (decentralized system, natural reuse of sub-systems in the hierarchy, event-driven nature of the computation  mitigating forgetting), there are still many open questions that need to be addressed.

\bibliography{main}{}
\bibliographystyle{plain}

\appendices
\section{A Detailed Description of the Architecture} \label{s:detailedDescription}

This appendix describes the various mechanisms of the ToyArchitecture Experts and how hierarchies of them interact. 
We  will first focus on describing the passive Expert which does not actively influence its environment, and is without context. Then, we will show how it can be extended with context (\ref{s:externalContext}) and actions (\ref{s:actions}). Afterwards, we will extend the definition of the context to allow experts in higher levels to send goals to the experts in lower levels (\ref{s:goalDirected}). We will define the exploration mechanisms (\ref{s:exploration}), and describe how a Reinforcement Learning (RL) signal can interface with the architecture so that it can learn from its actions. Together, these mechanisms implement distributed hierarchical goal-directed behavior.

\begin{table}[h!]
  \centering
	\begin{tabular}{ l l }
  	Variable & Description \\
  	\hline
    $T_h$							& Length of the past \\
    $T_b = T_h + 1$                 & Length of the lookbehind part \\
                                        &(past + current step)\\
    $T_f$							& Length of the future\\
    $m=T_b + T_f = T_h+1+T_f$		& Length of the whole sequence\\
  	$l \in 1, \dotsc, L$ 			& Index of a layer\\
  	$j \in J_l$ 					& Index of an Expert in the layer $l$\\
  	$V = \left\{ v_1, v_2, \ldots, v_K\right\}$ & Set of cluster centers (states)\\
  	                                    &of an Expert\\
    $D(\vec{x}) = |V| = K$ 			& Dimension of $\vec{x}$, number of\\
                                        &cluster centers\\
  	$\vec{O} = \vec{o}(1), \vec{o}(2), \ldots, \vec{o}(T)$ & Sequence of complete observations \\
  	$\vec{O}_j^l = \vec{o}_j^l(1), \ldots, \vec{o}_j^l(T)$ & Sequence of observations of \\
  	                                    &Expert $j$ in layer $l$ \\
    $\vec{x}_j^l(t)$ 				& Hidden state of the Expert $j$ in layer $l$ \\
    $\vec{y}_j^l(t)$ 				& Output vector of the Expert $j$ in layer $l$ \\
    $M = |S|$                       & Number of sequences considered in a TP \\
    $\mathcal{P}$                   & Set of all providers of context to\\
                                        &an Expert\\
    $S_c$                           & Likelihoods of seeing each context\\
                                        &element from each provider in each\\
                                        &position of each sequence\\
    \hline
\end{tabular}
\label{t:notation}
\caption{Selected notation.}
\end{table}

During inference, the task of an Expert $j$ in layer $l$ ($E_j^l$) is to convert a sequence of observations perceived in its own receptive field $\vec{o}_j^l(t)$ into a sequence of output values $\vec{y}_j^l(t)$. For simplicity, when discussing a single Expert, we will omit the $j$ and $l$ from the notation of the hidden states, observations, outputs, etc.

\subsection{The Passive Model without Context}\label{s:passive}

As discussed in \cref{s:resultingRequirements}, the process is split into the \textbf{Spatial Pooler}, the \textbf{Temporal Pooler}\footnote{The terminology adopted from~\cite{Hawkins2015}}, and \textbf{Output Projection}, which can be expressed by the following three equations:
\begin{align}
\vec{x}(t) & = f_1(\vec{o}(t) \vert \vec{\theta}_\text{SP}), \label{eq:1}\\
P(S)(t)    & = f_2(\vec{x}(t), \vec{x}(t-1), \ldots, \vec{x}(t-T_h) \vert \vec{\theta}_\text{TP}), \label{eq:2}\\
\vec{y}(t) & = f_3(P(S)(t)), \label{eq:3}
\end{align}
where the $\vec{\theta}_\text{SP}$ and $\vec{\theta}_\text{TP}$ are learned parameters of the  model.  
\subsubsection{Spatial Pooler}
The non-linear observation function from \cref{eq:1} is implemented by k-means\footnote{Due to simplicity and interpretability reasons, as described in \cref{s:motivation}.} clustering and produces one-hot vector over the hidden states:
\begin{equation} \label{e:dist}
    \vec{x}(t) = f_1(\vec{o}(t) \vert \vec{\theta}_\text{SP}) = \delta \left( \argmin_{v_i \in V} \left( \dist(v_i, \vec{o}(t)) \right) \right),
\end{equation}
where $\dist(\vec{v}_1,\vec{v}_2)$ is the $L^2$ Euclidean distance between two vectors $\vec{v}_1,\vec{v}_2$, and $V$ is a set of learned cluster centers of the Expert corresponding to the parameter $\vec{\theta}_\text{SP}$), and $\delta(v_i \in V)$ is a winner-takes-all (WTA) function which returns a one-hot representation of $v_i$. The observation function $f_1$ considers only the current observation and covers step number two (compression) as described in  \cref{s:resultingRequirements}. Separation is performed on a level of multiple Experts and is described in Appendix \ref{a:disentangled}.

Because we are learning from a stream of data, it might happen that some cluster centers in the Spatial Pooler do not have any data points and thus would be never adapted. There can be two underlying reasons for this: 1) the cluster centers were initialized far from any meaningful input, or 2) the agent has not seen some types of inputs for a long period (e.g. it stays inside a building for some time and does not see any trees). In situation 1, we would like to move the cluster center to an area where it would be more useful, but in situation 2, we typically want to keep the cluster center at its current position in order to not forget what was learned and have it be useful again in the future. We solve this dilemma by implementing a \textit{boosting algorithm} similar to \cite{Hawkins2006}. We define a hyper-parameter $b$ (\textit{boosting threshold}) and every cluster center, which has not received any data for the last $b$ steps, starts to be boosted  where it is moved towards the cluster center with highest variance among its data points. Using this parameter, we can modify the trade-off between adapting to new knowledge and not-forgetting old knowledge.     

\subsubsection{Temporal Pooler} \label{s:TemporalPooler}

The goal of the Temporal Pooler is to take into account a past sequence of hidden states $\vec{x}(0), \ldots, \vec{x}(t)$ and predict a sequence of future states $\vec{x}(t+1), \ldots, \vec{x}(t+T_f)$. Since the sequence of observations $\vec{O}_j^l$ might not have the Markovian property, and it might have been further compromised by the Spatial Pooler, the problem is not solvable in general. So we limit the learning and inference in one Expert to Markov chains of low order and learn the probabilities:
\begin{equation}
    P(\vec{x}(t+T_f),\dots,\vec{x}(t+1) | \vec{x}(t), \vec{x}(t-1), \ldots, \vec{x}(t-T_h)),
\end{equation}
which we express in the form of sequences $s_i = v_{i_1},  \ldots, v_{i_{T_h}}, v_{i_{T_h+1}}, v_{i_{T_h+2}} \ldots, v_{i_{m=T_h+1+T_f}}$. Each sequence can thus be divided into three parts of fixed size a history: $v_{i_1}, \ldots, v_{i_{T_h}}, v_{i_{T_h}}$, the current state: $v_{i_{T_h}+1}$ and a \textit{lookahead} part: $v_{i_{T_h+2}}, \ldots, v_{i_m}$, the entire sequence having the length of $m=T_h+1+T_f$. We call the history together with the current step the \textit{lookbehind} which is a sequence of length $T_b = T_h + 1$. See the bottom of \cref{f:example} for an illustration.

The theoretical number of possible sequences of hidden states grows very quickly with the number of states and the required order of Markov chains. But the observed sub-generator usually generates only a very small subset of these sequences in practice. Using a reasonable number of states and length of sequences (e.g. $N=30$ and $m=4$), it is possible to learn the transition model by storing all encountered sequences in each Expert $\Seq_j^l = \left\{ s_1, s_2, \ldots, s_M \right\}$ and computing their prior probabilities based on how often they were encountered. Then, the probability of the $i$-th sequence $P(s_i)$ is computed as:
\begin{equation} \label{e:normalizedSeqProb}
    P(s_i) = \Normp{\bar{P}(s_i)} =  \frac{\bar{P}(s_i)}{\sum_{s_j \in \Seq}\bar{P}(s_j)},
\end{equation}
where $\Normp{D}$ denotes the normalization of values $D$ to probabilities, and:
\begin{equation} \label{e:sequenceProbNonnormalized}
    \bar{P}(s_j) = P_{pr}(s_j) \cdot P(s_j | \vec{x}(t-T_h:t)),
\end{equation}
where $\Prior(s_j)$ is the prior probability of the sequence $s_j \in \Seq$ (i.e. how often it was observed relative to other sequences), $P(s_j | \vec{x}(t-T_h:t) = \prod_{d = 0}^{T_h} \Iseq(s_j, d, \vec{x}(t-T_h+d))$ is the match of the beginning of the sequence $s_j$ with the recent history of states $\vec{x}(t-T_h), \dots , \vec{x}(t)$, and $\Iseq(s_j, d, \vec{x}) \mapsto \left\{ 1 - \epsilon, \epsilon \right\}$ is an indicator function producing a value close to $1$ if the hidden state $\vec{x}$ corresponds to the cluster $v_{j_d}$ at the $d$-th position in the sequence $s_j$, otherwise $\epsilon$\footnote{$\epsilon > 0$ is a small constant ensuring each sequence has a nonzero probability and that sequences corresponding at least partially to the data  have higher probabilities than those which do not correspond at all.}. The parameter $T_h < m$ defines the fixed length of the required match of the sequence, so given $T_h=2$ and $m=5$, the sequence probabilities will be computed based on the first $T_h+1=3$ clusters in the sequence. Sequences such as this can be used for predicting 2 steps into the future. The value $P(S)(t)$ from \cref{eq:2} is then a probability distribution over all sequences in time step $t$:
\begin{equation}
P(S)(t) = \left\{ P(s_i)(t):  s_i \in \Seq \right\}.
\end{equation}
These are the main principles behind learning and inference of the Temporal Pooler.

\subsubsection{Output Projection}

Finally, the Expert has to apply the output function described in \cref{eq:3}. In each time step, the output function takes the the current sequence probabilities and produces the output of the Expert $y(t)$:
\begin{equation} \label{e:outputFunc}
    \vec{y}(t) = f_3(\left\{ P(s_i)(t): s_i \in \Seq \right\}).
\end{equation}
When defining the output function, the following facts need to be taken into account: the outputs $\vec{y}_j^l$ of Experts in layer $l$ are processed by the Spatial Poolers of Experts in  layer $l+1$ where the observations are clustered based on some distance metric. There are two extreme situations:
\begin{itemize}
\item In the case where the sequence of states $\vec{x}_j^l(t)$ for the child Experts $j \in J_l$ on layer $l$ are not predictable, the parent Experts in layer $l+1$ should form their clusters mostly based on the spatial similarities of the hidden states of the Experts\footnote{The one-hot output of the Spatial Pooler does not fulfill this requirement. But the spatial similarity is preserved over the outputs of multiple Experts which receive similar inputs (distributed representation). The parent Experts $E_i^{l+1}$ then receive outputs of multiple experts from layer $l$, therefore they perceive the code which preserves the spatial similarity.} in layer $l$. This way, the details of the unpredictable processes are preserved as much as possible and passed into higher layers of abstraction where these uncertainties can be resolved.
\item On the other hand, in the case where the state sequences ${\vec{x}_j^l(t)}_t$ are perfectly predictable, the spatial properties of the observations are relatively less important than their behavior in time, and the clustering in layer $l+1$ should be performed based on the similarities between sequences (i.e. temporal similarity).
\end{itemize}

Based on these properties, the output function should be defined so that the resulting hierarchy implements \textbf{implicit efficient data-driven allocation of resources}. The parts of the process that are easily predictable by separate Experts low in the hierarchy will be compressed early. The unpredictable parts of the process will propagate higher into the hierarchy where the Experts try to predict them with more abstract spatial and temporal contexts. This is a compromise between sending what the architecture knows well vs just sending residual errors~\cite{Spratling2017}. 

In the current version, we use the following output projection function: The output dimension $D(\vec{y})$ is fixed to the same number of hidden states in the Expert, and $D(\vec{x}) = |V| = K =D(\vec{y})$. The output function is defined as follows:
\begin{multline} \label{e:outputComputation}
    \vec{y}_j^l(t) = f_3(P(s_i\in \Seq_j^l)(t)) = \\
\Normp{\vec{x}(t) + \sum_{k \in 1 \isep K} \delta({x_k}) \sum_{i \in 1 \isep M} \bar{P}(s_i) \sum_{d \in 1 \isep m } \Iseq(s_i, d, \delta(x_k))},
\end{multline}
where $I$ is the indicator function from \cref{e:sequenceProbNonnormalized}, $\delta$ is the WTA function from \cref{e:dist}, $\Normp{.}$ is the normalization function from \cref{e:normalizedSeqProb}, and $\bar{P}(s_i)$ is the probability that we are currently in sequence $s_i$ from \cref{e:sequenceProbNonnormalized}. This definition of the output function has the following properties:
\begin{itemize}
\item In the case that the observation sequence is not predictable, the predictions from sequences with high probability will have high dispersion over future clusters. Therefore the position corresponding to the current hidden state and recent history of length $T_h+1$ will be dominant in the output vector $y(t)$. So the parent Expert(s) in layer $l+1$ will tend to cluster these outputs mostly based on the recent history of length $T_h+1$ as opposed to the predictions.

\item In the case that the observation sequence is perfectly predictable, only one sequence will have high probability in each time step, so both the past and predicted states will have high probability. Therefore the parent Expert(s) in $l+1$ will tend to cluster based on the predicted future more than in the previous case. The sequence of observations for $E_i^{l+1}$ will therefore be more linear (similar to a sliding window over the recent history and future), therefore it will be possible to chunk the observations more efficiently. More importantly: the output of these parent Experts $y_i^{l+1}(t)$ will correspond more to the future (since the lower-level Experts are predicting better). As a result, the higher levels in the hierarchy should compute with data which correspond to the increasingly more distant future. This way the hierarchy does not think about what happened but rather what is going to happen.

\end{itemize}

This means that the temporal resolution in higher layers is determined automatically based on the predictability of the observations in the current layer, and this resolution can dynamically change in time. Since the clustering applies a strict winner-takes-all (WTA) function, and the Temporal Pooler does not accept repeating inputs, the entire mechanism naturally results in a \textbf{completely event-driven architecture}.

\subsection{The Passive Model with External Context}\label{s:externalContext}

Until now, the goal of each Expert has been to learn a model of the part of the environment solely based upon its own observations $\vec{o}_j^l$. This can lead to highly suboptimal results in practice. Often it is necessary to use some longer-term spatial or temporal dependencies as described in \cref{s:resultingRequirements}. A context input (see \cref{f:expert}) is used to provide this information.

The meaning and use of the bottom-up and context connections should be asymmetrical: the bottom-up (excitatory) connections decode \qq{visual appearance} of the concept, while top-down (modulatory) connections~\cite{Adams} help to resolve the interpretation of the perceived input using the context. This asymmetry should prevent  positive feedback loops in which the bottom-up input might be completely ignored during both learning and inference. As a result, the hidden state of the architecture should still track actual sensory inputs.

The context input can be then seen as a \textbf{high-level description of the current situation} from the Expert's surroundings (both from the higher level and possibly from neighboring experts in the same layer).

\begin{figure}[ht]
    \centering
    \includegraphics[width=\linewidth]{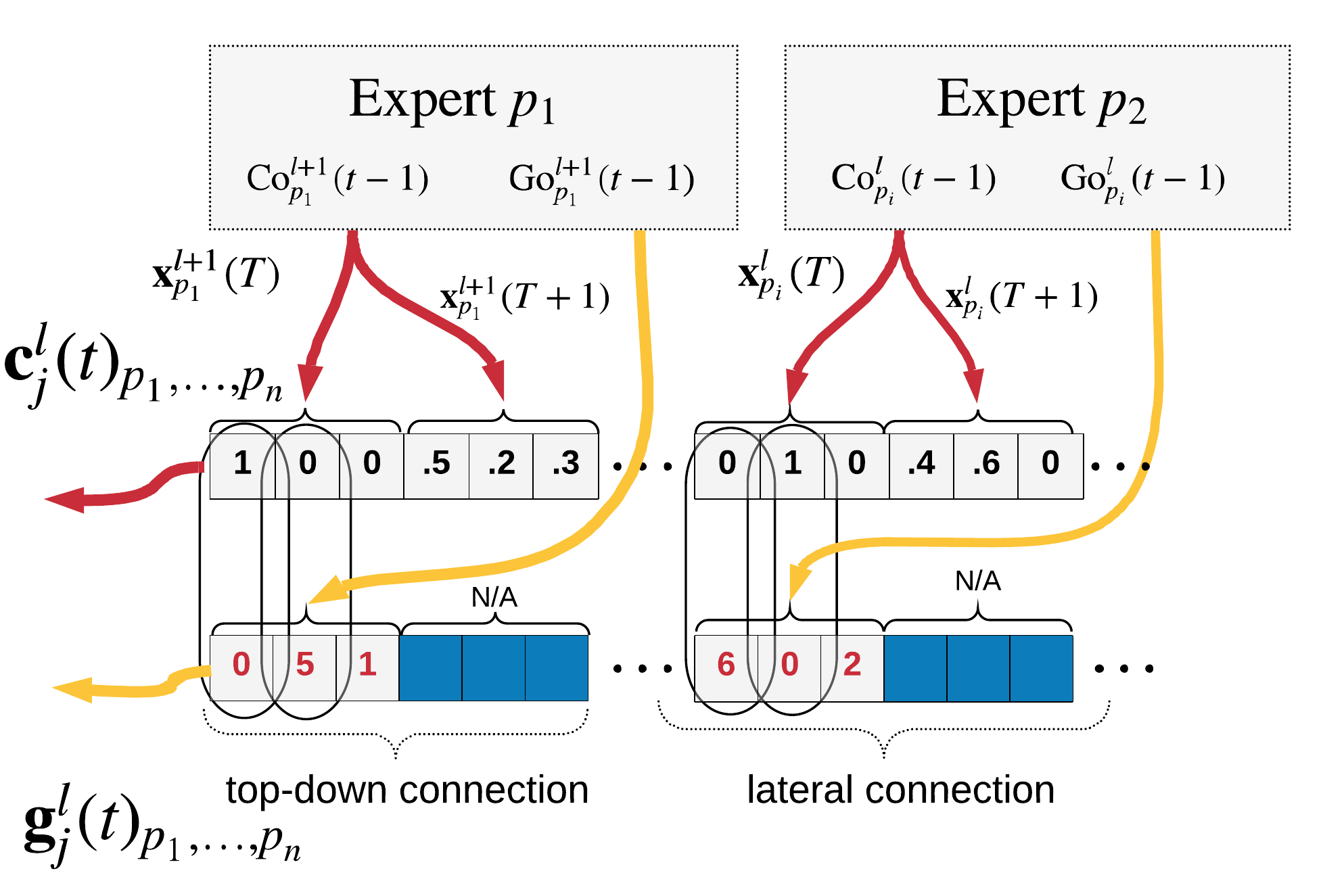}

    \caption{Context and Goal input vectors to Expert $E_j^l$. Both are collections of top-down and lateral inputs from other Experts from the previous time step. The Goal input has some parts masked-out (blue parts). The resulting two input vectors can be interpreted as a high-level description of the current state, and a passive prediction of what will happen next ($\vec{x}_j^l(T)$, $\vec{x}_j^l(T+1)$), and the goal as a preference (measured in the expected value of reward) for the next state. Note that in this figure, the variable $t$ denotes the time for the Expert receiving the context ($E_j^l$), while $T$ denotes the time for an Expert sending the context. Because all Experts are event driven, the time between two changes in an Expert states is different for different Experts.}
    \label{f:contextV2}
    
\end{figure}

It is possible to use various sources of information as a context vector, such as:
\begin{itemize}
\item \textbf{Past activity} of other Experts: this extends the ability of $E_j^l$ to take into account dependencies further in the past.
\item \textbf{Recent activity} of other Experts: this increases spatial and temporal range.
\item \textbf{Predicted activity} of other Experts: this extends the ability of $E_j^l$ to distinguish the recent observation history according to the future. This process could be likened to Epsilon Machines, where the idea is to differentiate histories according to their future impact~\cite{Tan,Brodu2011}.
\end{itemize}
The context output of an Expert: $Co_i^{l}(t) = \langle \vec{x}_j^l(t), \vec{x}_j^l(t+1) \rangle$ is a concatenation of the Spatial Pooler output (i.e. the winning cluster for this input) and the Temporal Pooler prediction of the next cluster. The goal $Go_i^{l}(t)$ is also attached to the context (see \cref{f:contextV2}), but we will talk about them separately for clarity. The ensemble can be thought of colloquially as communicating: \qq{Where, I am}, \qq{Where I expect to be in the future}, and \qq{What reward I expect for each possible future clusters}.

The context input is a collection of context outputs (refer to the red lines in the \cref{f:expert}) from multiple other Experts. Each Expert supplying context is known as a \textit{provider}, and there is no distinction between parent providers and lateral providers\footnote{Note that in general context connections that skip multiple layers are allowed as well.}. The context input to Expert $E_j^l$ is therefore defined as:  
\begin{equation} \label{e:contextConcat}
    \vec{c}_j^l(t) = (\langle {\vec{x}_i}_{(T)}(t-1), {\vec{x}_i}_{(T+1)}(t-1)  \rangle: i \in \mathcal{P}_j^l)
\end{equation}
where $\langle . \rangle$ denotes concatenation, $\mathcal{P}_j^l$ is set of providers for $E_j^l$, and ${\vec{x}_i}_{(T)}(t-1)$ and ${\vec{x}_i}_{(T+1)}(t-1)$ are the current ($T$) and predicted ($T+1$) clusters of provider $i$ from the previous step\footnote{The variable $t$ denotes the time for the Expert receiving the context ($E_j^l$), while $T$ denotes the time for an Expert sending the context (because all Experts are event driven, the time between two changes in an Expert states is different for different Experts).} respectively.

Context is incorporated in the Temporal Pooler prediction process by having the TP learn the likelihoods of each context element from each provider being 1, for each lookbehind cluster in each sequence as ${S_C}_j^l$. 

In using the context during inference, we augment the calculation of the unnormalised sequence probabilities (\cref{e:sequenceProbNonnormalized}) by also matching the current history of contexts $\vec{c}_j^l(t - T_h), \ldots, \vec{c}_j^l(t)$ with the remembered sequence contexts ${S_C}_j^l \in \theta_{TP}$.

We start by extending the definition of $P(S)(t)$ in \cref{eq:2}: 
\begin{align}
    P(S)(t) =& f_2\big(\vec{x}(t), \ldots, \vec{x}(t - T_h), \nonumber \\
    & \vec{c}(t), \ldots, \vec{c}(t - T_h)| \theta_{TP}\big) 
\end{align}
We consider each context provider separately. For each sequence, we calculate the likelihood of that sequence based on the history from each individual provider $p \in \mathcal{P}$:
\begin{equation}
    P_c(S_j^l)(p) = \vec{c_j^l}(t - T_h:t)(p) \cdot {S_C}_j^l(p)
\end{equation}
Considering the role of the context, we wish that in a world where multiple sequences are equally probable, the context will disambiguate the situation. Given that ${S_C}_j^l$ is learned alongside $S_j^l$, in a situation where each Expert has the same data, the contexts should correlate highly with $\bar{P}(S)$ and the predictions based solely on the context history would be approximate to the predictions using the cluster history and priors:
\begin{equation}
    \bar{P}(S) \approx \mathbb{E}_{p_i}(P_c(S)(p_i))
\end{equation}
But in reality, each Expert might be looking at a different receptive field and have generally different information. On the other hand, context from most of the Experts can be of no use for the recipient and it is probable that it will be highly correlated among the providers. Thus averaging the predictions based on the individual contexts might obscure the valuable information. So rather than using every context equally for disambiguation, we would like to use only the most informative one. We choose the most unexpected context to use, as the context which is the most disruptive to the otherwise anticipated predictions is likely to contain the most information about the current state of the agent and environment. As a metric of unexpectedness, we use the Kullback-Leibler\cite{Kullback1951} divergence between the predictions based on the history of cluster centers and one \qq{informative} context vs predictions based just on the history of cluster centers. 

We therefore update \cref{e:normalizedSeqProb} to include this selection and use of the most informative context:
\begin{align}
    P(s_i) = \langle \bar{P}(s_i) &\cdot \mathcal{F}(\bar{P}(s_i), P_c(s_i)) \rangle_1 \text{, where} \label{e:normalizedSeqProb2}\\
    \mathcal{F}(\bar{P}(s_i), P_c(s_i)) &= {S_C}(\argmax_{p\in\mathcal{P}}  D_{KL}(\langle \bar{P}(s_i) \rangle_1 || \nonumber \\
     & \quad \langle \bar{P}(s_i) \cdot P_c(s_i) \rangle_1))
\end{align}
As a result, using the context as a high level description of the current situation, each Expert can also consider longer spatial and temporal dependencies which defy the strict hierarchical structure (see \cref{s:predominantlyHierarchical}) in order to learn the model of its own observations more accurately.

\subsection{Actions as Predictions}\label{s:actions}

Until now, the architecture has been only able to passively observe the environment and learn its model. Now, the mechanisms necessary to actively interact with the environment (i.e. to produce actions) will be introduced with as small a change to the architecture as possible. 

From the theoretical perspective, the HMM can be extended to a Partially Observable Markov Decision Process (POMDP)~\cite{Sutton}. While remembering that each Expert processes Markov chains of order $m-1$, the decision process corresponds to the setting:
\begin{multline} \label{e:nonmdp}
P\big(\vec{x}(t) | \vec{x}(t-1), a(t-1), \ldots , \vec{x}(0),a(0)\big) \\
= P\big(\vec{x}(t)|\vec{x}(t-1),a(t-1), \ldots , \vec{x}(t-m),s(t-m)\big),
\end{multline}
where $a(t)$ denotes an action taken by the Expert at time $t$. Note that this setting could be treated as a task for active inference, where the agent proactively tries to discover the true state of the environment if necessary~\cite{Whitehead1995a, Friston2009}. But for now, we will consider a similar approximation of the problem as in the previous sections and leave and explicit active inference implementation to future work.

Since we want the hierarchy to be as general as possible, it is desirable to define the actions in such a way that they can be used in case the Expert has the ability to control the actuators (either directly or indirectly through other Experts), but do so that they will not harm the performance in the case that the Expert is not able to perform actions, and can only passively observe.

\begin{figure}[ht]
    \centering
    \includegraphics[width=\linewidth]{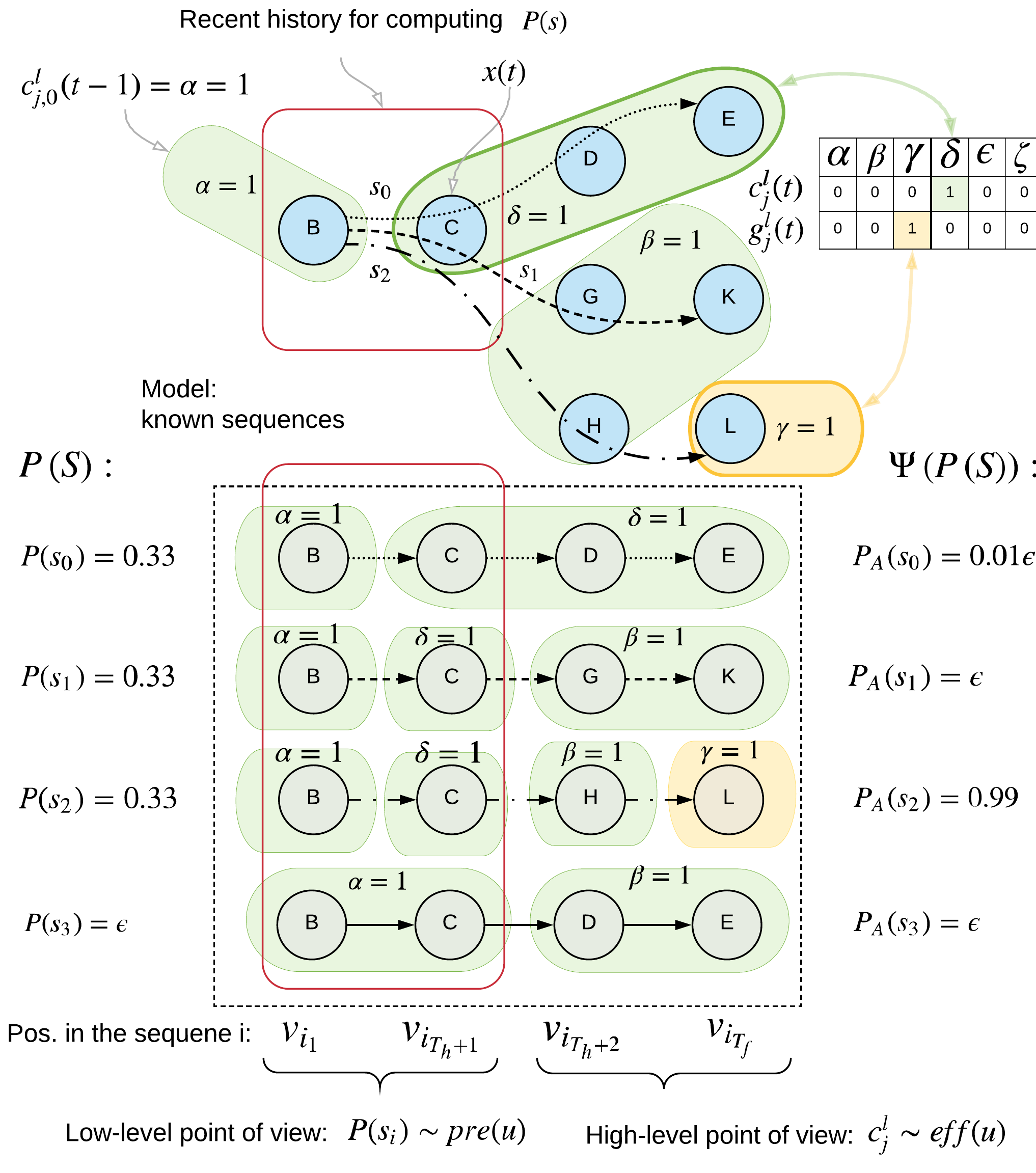}
    
    \caption{An example of a recent sequence of states (where $T_h=1$, $T_f=2$) $\vec{x}(t-T_h:t)$ in $E_j^l$ which shows: 1) The sequence of context inputs $\vec{c}_j^l(t-T_h:t)$ helping to resolve uncertainty during computation of $P(s)$; 2) A goal vector $\vec{g}_j^l(t)$ defining the the expected rewards of the target state \textbf{Bottom}: the library of learned sequences $\Seq_j^l$, each sequence is defined by an ordered list of states $x$, each potentially in a different context $c_j^l$. \textbf{Top}: visualization of the current state $\vec{x}(t)$ and several possible futures. These futures are estimated based on the content of the Model. First, the probability distribution $P(s_i)$ is computed based on a sequence of recent states and contexts. Then, the sequence probabilities are increased proportionally to the probability that the reward can be obtained by following that sequence (updated sequence probabilities $P_A(s_i) = \Psi \left (P \left (\Seq \right ) \right )$ based on the reward are depicted on the right). As a result, this increases the probability of choosing the state $H$ as an action\textemdash setting it as a goal output $\Go_j^l(t)=H$. In this example, the first 3 sequences are equally matched by the $\vec{x}(t-T_h:t)$ and $\vec{c}(t-T_h:t)$ therefore they have equal probabilities. But after applying the $\Psi$, the $s_0$ has the smallest probability, the $s_1$ has higher probability since it sets the $\delta$ to zero in the future, but the $s_2$ has the highest probability, because it both: sets the $\delta$ to zero and $\gamma$ to one, as required by the goal input $\vec{g}_j^l(t)$.}
      
    \label{f:example}
\end{figure}

For this reason, actions are not explicitly modeled in this architecture. Instead, \textbf{an action is defined as an actively selected prediction of the next desired state}\footnote{This actively chosen predicted state action should be reachable from the current state $\vec{x}(t)$ with a high probability, i.e., be in coherence with what is possible.} $\vec{x}(t+1)$. The selected action (desired state in the next step) is indicated on the Goal output of the Expert (see \cref{f:expert}): $Go_j^l(t)$.

Given the library of sequences $S$, the recent history of hidden states $\vec{x}(t-T_h:t)$, and the context inputs $\vec{c}(t-T_h:t)$, the Expert computes the sequence probabilities $P(S)$ using \cref{e:normalizedSeqProb2}. Then, those sequence probabilities are altered based on preferences over the states to which they lead (see Appendix \ref{s:goalDirected} and \ref{s:rl}). This results in a new probability distribution
\begin{equation}
	P_A(S) = \Psi \big ( P(S) \big ),
    \label{e:psi}
\end{equation}
where $\Psi$ can be seen as a sequence selection function, see \cref{f:example} for illustration. Finally, the Goal output of the Expert for the next simulation step $\Go_j^l(t)$ is computed (see \cref{f:expert}). This can be seen as actively predicting the next position in a sequence:
\begin{equation} \label{e:asm}
	\Go_j^l(t) = \Theta \left ( \Phi (P_A(S)) \right ),
\end{equation}
where $\Phi$ converts the probabilities of sequences $P(\Seq)(t)$ into a probability distribution over clusters $P(V)(t+1)$ predicted in the next step:
\begin{equation} \label{e:Phi}
P(v \in V) = \Phi \left (P(s \in S) \right ) = \sum_{s \in \Seq} P(s) \cdot \Iseq(s, T_h+2, \delta(v)),
\end{equation}
where $I$ is the indicator function from \cref{e:sequenceProbNonnormalized}, position $T_h+2$ corresponds to the next immediate step and $\delta$ is the WTA function from \cref{e:dist}. The $\Theta$ in \cref{e:asm} is an action selection function for which it is possible to use multiple functions, namely identity, $\epsilon$-greedy selection, sampling, or $\epsilon$-greedy sampling. 

In the example in \cref{f:example}, without considering any preferences over the sequences (the $\Psi$ in \cref{e:psi} collapses to identity), the probabilities of the first three sequences are equal, therefore the function $\Theta$ would choose the states $D$, $G$ and $H$ with equal probability.

The whole process can be seen as follows: Each Expert throughout the hierarchy, calculates a plan based on a short time horizon $T_f$, chooses the desired imminent actions (states one step in the future which are desired and probably reachable) and encodes this information as the Goal output $\vec{g}_j^l$. This signal is then either received by other Experts and interpreted as the goal they should reach), or used directly by the motor system in the case that the Expert is able to control something.

In the presented prototype implementation, the desirability of the goal states is encoded as a vector of rewards that the parent expects that the architecture will receive if the child can produce a projection $\vec{y}(t+1)$ which will cause the parent SP to produce the hidden state $\vec{x}(t+1)$ corresponding to the index of the goal value.

An expert receiving a goal context computes the likelihood of the parent getting to each hidden state using its knowledge of where it presently is (\cref{e:normalizedSeqProb2}), which sequences will bring about the desired change in the parent (\cref{e:psi2}), and how much it can influence its observation in the next step $\vec{o}(t+1)$ by its own actions (see Appendix \ref{s:exploration}). It rescales the promised rewards using these factors, combines them with knowledge about its own rewards (see Appendix \ref{s:rl}) and then calculates which hidden states in the next step correspond to sequences leading towards these combined rewards. From here, it either publishes its own goal $Go$ (expected reward for getting into each cluster), or if it interacts directly with the environment picks an action to follow\footnote{The action of bottom level Experts at $t-1$ is provided on $\vec{o}(t)$ from the environment, so the picking of an action is equivalent to taking the cluster center of the desired state and sampling the actions from the remembered observation(s).}. This mechanism is described in more details in the following section.

\subsection{Goal-directed Inference} \label{s:goalDirected}

This section will describe the mechanisms which enable the Expert:

\begin{itemize}

\item To decode the goal state received from an external source (usually other Experts).
\item To determine to what extent the goal state can be reached, or at least if the distance between the current state and the goal can be decreased. 
\item To make a first step (\qq{action}) leading towards this goal if it is possible by setting $\Go_j^l$ to an appropriate value.

\end{itemize}

As a result, these mechanisms should allow the hierarchy of Experts to act deliberately. The architecture will hierarchically decompose a decision\textemdash potentially a complex plan, represented as one or several steps on an abstract level, into a hierarchy of short trajectories. This corresponds to the ability to do decentralized goal-directed inference, which is similar to hierarchical planning (e.g. state-based Hierarchical Task Network (HTN) planning~\cite{Georgievski2014}). Note that such a hierarchical decomposition of a plan has many benefits, such as the ability to follow a complex abstract plan for longer periods of time, but still be able to reactively adapt to unexpected situations at the lower levels. There are also theories that such  mechanisms are implemented in the cortex \cite{Pezzulo2018}.

In this section, we will show a simple mechanism which approximates the behavior of a symbolic planner. This demonstrates one important aspect: the hierarchy of Experts converts the input data into more structured representations. On each level of the hierarchy the representation can be interpreted either sub-symbolically or symbolically. This gives us the ability to define \textbf{symbolic inference mechanisms on all levels} of the hierarchy (e.g. planning), which then \textbf{use grounded representations}.

Furthermore, in Appendix \ref{s:rl}, we will show how a reinforcement signal can be used for setting preferences over the states in each Expert. This will in fact equip the architecture with model-based RL~\cite{Nagabandi}. It also means that \textbf{locally reachable goal states can emerge across the entire hierarchy} with them appearing on different time scales and levels of abstraction, which leads to completely decentralized decision making.

The main idea of goal-directed inference is loosely inspired by the principles of predictive coding in the cortex~\cite{Spratling2017}, where it is assumed that each region tries to minimize the difference between predicted and actual activity of neurons. In ToyArchitecture, a more explicit approach for determining the desired state is used. The approach can be likened to a simplified, propositional logic-based version~\cite{Kautz} of the symbolic planner called Stanford Research Institute Problem Solver (STRIPS)~\cite{Fikes}. In this architecture, each Expert will be able to implement forward state-space planning with a limited horizon~\cite{Ghallab}.

\textbf{STRIPS definition:~} Let $\mathcal{L}$ be a propositional language with finitely many predicate symbols, finitely many constant symbols, and no function symbols. A restricted state-transition system is a triple $\Sigma^p=(S^p,A^p,\gamma^p)$, which is described in \cref{t:strips}. 

\begin{table}[ht]
  \centering
\begin{tabular}{ l p{5cm} }
  variable & meaning \\
  \hline
  $s_i^p={x_1,x_1,..}$							& State\textemdash set of ground atoms of $L$\\
  $S^p=\{s_1^p,s_2^p,...\}$ 					& Set of states \\
  $U = \{u_1,u_2,...\}$ 						& Set of operators (actions)\\
  $\gamma: S^p \times U \rightarrow S^p$ 		& State transition function\\
  $u = (\pre(u), \eff(u))$						& Operator \textemdash transforms one state\\ 
  												& to another, if applicable\\
  $\pre(u) = \{\pre^+(u), \pre^-(u)\}$			& Precondition\textemdash set of literals which \\ 
  												& determines if the operator\\ 
  												& is applicable\\
  $\eff(u) = \{\eff^+(u), \eff^-(u)\}$			& Effect\textemdash set of literals which\\
                                                    &determines how the operator\\
                                                    &changes the state if applied\\
\hline

\end{tabular}
\caption{STRIPS language definition.}\label{t:strips}
\end{table}

State $s^p$ satisfies a set of ground literals $g^p$ (denoted $s^p \vDash g^p$) iff: every positive literal in $g^p$ is in $s^p$ and every negative literal $g^p$ is not in $s^p$. It is possible to represent states $s^p$ as binary vectors (where each ground literal corresponds to one position in the vector) and operators/actions $u$ as operations over these vectors.

The operator $u$ is applicable to the state $s^p$ under the following conditions. %
\begin{align}
\pre^+(u) & \subseteq s^p,\\
\pre^-(u) & \cap s^p = \{\}.
\end{align}
Then, the state transition function $\gamma$ for an applicable operator $u$ in state $s^p$ is defined as:
\begin{equation}
\gamma(s^p,u) = \large ( s^p-\eff^-(u^p) \large ) \cup \eff^+(u^p).
\end{equation}
The STRIPS planning problem instance is a triple $P^p=(\Sigma^p, I^p, G^p)$, where: $\Sigma^p$ is the restricted state-transition system described above, $I^p \in S^p$ is the current state and $G^p$ is a set of ground literals describing the goal state (which means that $G^p$ describes only required properties, which are a subset of the propositional language $\mathcal{L}$). 

Given the planning instance $P^p$, the task is to find a sequence of operators (actions), which consecutively transform the initial state $I^p$ into a form which fulfills the conditions of the goal state $G^p$.

One possible method to find such a sequence is to search through the state-space representation. Since the decision tree has potentially high branching factor, it is useful to apply some heuristic while choosing the operators to be applied. To quote from the original paper~\cite{Fikes}: \qq{We have adopted the General Problem Solver strategy of extracting differences between the present world model [state] and the goal, and of identifying operators that are relevant to reducing these differences}.

Now we will describe how an approximation of this mechanism is implemented in ToyArchitecture.

\textbf{Similar mechanisms in the ToyArchitecture:}~ The architecture learns sequences $s_i$ of length $m$ and each step in the sequence corresponds to an action. Each sequence is a trajectory in the state-space of Expert $E_j^l$ (see states with big letters and transitions between them in the \cref{f:example}). But, more crucially, from the point of view of the parent\footnote{For simplification, we can consider one parent Expert, but the approach generalizes to top-down connections from multiple parents as well as multiple lateral connections from other Experts in the same layer simultaneously.} Expert $E_j^{l+1}$, \textbf{each sequence can be seen as an operator $u$}.

For the purposes of planning, aside the context vector input $\vec{c}_j^l$, the Expert is equipped with a goal vector input $\vec{g}_j^l$, which specifies the goal description $G$. 

From the point of view of $E_j^{l}$, $\vec{c}_j^l(t)$ describes the current state (corresponds to $s_i^p(t)$ in the STRIPS), while $\vec{g}_j^l(t)$ describes a superposition of desirable goal states. With each position marked by a real number indicating how preferable the state is for the parent\footnote{We can think of this as the expected value of the state for the parent.}.

Note that (as explained in Appendix \ref{s:externalContext}) each Expert learns the probabilities of sequences dependent on context and position in the sequence (\cref{e:normalizedSeqProb2}) and stores them the form of a table of frequencies of observations of each combination. This allows us to define the operator $u_i=\{\pre(u_i), \eff(u_i)\}$ (corresponding to the learned sequence $s_i$) in a stochastic form, where we define the probability of $\eff(u_i)$ being $\vec{\hat{c}}$ as the probability that the ending clusters of the sequence $s_i$ will be observed in the context $\vec{\hat{c}}$:
\begin{equation} \label{e:eff}
P\left( \eff(u_i) = \vec{\hat{c}} \right) = \max_{f \in 1 \isep T_f} P\left(\vec{c}_j^l(t+f) = \vec{\hat{c}} \vert s_i(t)\right).
\end{equation}
The precondition $\pre(u_i)$ determining the applicability of the operator can be also defined in a stochastic manner as the probability that the Expert is currently in the sequence $s_i$:
\begin{equation} \label{e:pre} 
	P\left( \pre(u_i) = \vec{c}_j^l \right) = P^G\left (s_i(t) \right),
\end{equation}
where $P^G\left (s_i(t) \right)$ is a probability of the sequence $s_i$ similar to \cref{e:normalizedSeqProb2}, but computed for the situation when the Expert actively tries to influence it (see Appendix \ref{s:exploration} for more details). Note that \cref{e:eff,e:pre} imply that the meaning of the operators $u_i=\{\pre(u_i), \eff(u_i)\}$ is different in each Expert and each time step.

Finally, the sequence selection function $\Psi$ from \cref{e:psi} can be defined as follows:
\begin{multline} \label{e:psi2} 
	\Psi\left(P(s_i(t))\right) =\\
    \Normp{P\left( \pre(u_i) = \vec{c}_j^l(t) \right) \cdot P\left( \eff(u_i) = \vec{g}_j^l(t) \right)} = \\
    \Normp{P^G(s_i(t)) \cdot \max_{f \in 1 \isep T_f} P\left(\vec{c}_j^l(t+f) = \vec{g}_j^l(t) \vert s_i(t)\right)},
\end{multline}
where $\Normp{.}$ denotes normalization to probabilities described in \cref{e:normalizedSeqProb}.

This means that each Expert can implement deliberate decision making, looking ahead $T_f$ steps into the future. Each step, it looks for currently probable sequences which maximise the expected value when moving the parent from the current context vector $\vec{c}_j^l(t)$ to the state dictated by $\vec{g}_j^l(t)$.

\begin{algorithm} \label{a:asm}
 \KwData{
 	Observation history $\vec{o}_j^l(t-T_h:t)$, \\
 	Context history $\vec{c}_j^l(t-T_h:t)$, \\
    Goal description $\vec{g}_j^l(t)$}
 \KwResult{Goal output $Go_j^l(t)$ }
 Compute applicability of the operators: compute sequence probabilities $P(s_i)(t)$ (\cref{e:pre})\;
 
 Select operators that are applicable and have high chance of achieving one of the the goal states: weight sequence probabilities by these $\Psi(P(s_i)(t))$ (\cref{e:psi2})\;
    
 Compute the probabilities of preferred states in the next step $\vec{x}(t+1)$ by $\Phi(\Psi(P(s_i)(t)))$ (\cref{e:Phi})\;
 
 \eIf{Expert is to produce an action}
 {Apply an action selection function $\Theta$ (\cref{e:asm})\;
 Set the selected action to the $Go_j^l(t)$\;}
 {Set $Go_j^l(t)$ to the values of the received expected values weighted by the computed next step probabilities\;}
 
 \caption{Goal directed inference\textemdash an approximation of a stochastic version of STRIPS state-space planning with a limited horizon. Describes how the Expert decides on which action to apply in order to maximise the expected value of rewards communicated in $\vec{g}_j^l(t)$ from the current context $\vec{c}_j^l(t)$. If an Expert is directly connected to the actuators, then an action is selected directly, otherwise the expert propagates the expected values of the states to its children.}
\end{algorithm}

The entire process is summarized in Algorithm \ref{a:asm} and illustrated on an example in \cref{f:example}. Compared to STRIPS, each Expert can plan with only limited lookahead, but this decision is decomposed into sub-goal of Experts in lower layer. This leads to the efficient hierarchical decomposition of tasks. Moreover, compared to classical symbolic planners, \textbf{representations in the hierarchy are completely learned from data} and since the Experts still compute with probabilities, the inference is stochastic and can be interpreted as continuous and sub-symbolic.

\subsection{Reinforcement Learning}\label{s:rl}

In the previous section we described how the Expert can actively follow an externally given goal. The same mechanism can be used for reinforcement learning with a reward $r \in \mathbb{R}$.

When reaching a reward, every Expert in the architecture gets the full reward or punishment value. During learning, each Expert assumes that it was at least partially responsible for gaining the reward and therefore associates the reward gained at $t$ with the state/action pair at $t-1$, so that for all $\{(\vec{x}(t), a)|, a \in A_j^l\}$ there is a corresponding $r \in {S_R}_j^l$ which is an estimate of the reward gained when in state $\vec{x}$ and taking action $a$. Because the Experts are event driven, they sum up all the rewards received during the steps they did not run (their cluster did not change).

The initial expert reward calculation is: 
\begin{equation}
    \mathbb{E}[R_j^l(t \ldots T_f)] = Go_j^l(t) \cdot {S_C}_j^l(t \ldots T_f) + {S_R}_j^l(t \ldots T_f)  
\end{equation}
This is the expected value of the promised reward from each provider, for each future state in each sequence. Any rewards that the Expert can `see' from this point are also included as the term ${S_R}_j^l(t \ldots T_f)$). 

The action which the Expert should perform is related to the sequence that it wants to move to. As it is trying to maximise its rewards, the Expert should pick an action which would position it in a  which has the highest likelihood of obtaining the most rewards. As this is the expected value (and also assume that rewards are sparse, and that an Expert can only expect reward once in the current lookahead (i.e. $T_f$) of the sequence.), the maximum of rewards from the sequence is used\footnote{Taking the maximum as a lower bound on the expected reward works only in case rewards are all non-negative or all non-positive. In case we want the agent to accept both rewards and punishments at once, they need to be processed separately and combined just in the lower Experts sending the actual actions to the environment.}:
\begin{equation}
    \mathbb{E}[R(t+1)] = \max_{f \in 1 \ldots T_f} R(t + f) \cdot I(S)(t+f) \cdot P(S)(t) \cdot d^f
\end{equation}
where $I(S)$ is the influence model, which is a model of how able the Expert is to move from one state to another by taking an action (see Appendix \ref{s:exploration} for how this is calculated) $d^f$ is a discount factor through time and $P(S)(t)$ is the probability distribution over sequences that the Expert is in at time $t$.

From the perspective of the Expert, $r_i = \argmax_i \mathbb{E}[R(t+1)]$ is the best possible reward, $s_i(t+1)$ is the best possible sequence, and $a(t)$ is the best possible action for the Expert to take given the probability of the current state, promised rewards, and the probability of affecting future states. The action $a(t)$ is therefore the one taken by the Expert if it is expected to interact with the outside world. Otherwise the output goal of this layer $Go_j^{l-1}(t+1)$ is set to the values of $\mathbb{E}[R(t+1)]$ thus propagating a promise of the expected reward to to its children.

Because goal-directed inference is hierarchically distributed through the whole architecture, the external reward $r$ has to be provided to each Expert\footnote{Any type of reward decomposition~\cite{Dietterich1998} would be subject of a future work.}. In this way, a top level Expert tries to get to an abstract state where the agent received a reward (for example a quadrant of a map), where the rewarding state is already reachable by some middle layer expert (within its finite horizon $T_f$) and the agent is driven closer to the reward with higher precision (e.g., to a particular room) until finally a low-layer expert is able to find a sequence of atomic actions leading precisely to the rewarded state.

Thus the standard way of dealing with long term rewards by artificially distributing them along traces~\cite{Sutton} can be completely avoided in the case of hierarchically distributed goal-directed inference with enough granularity, because there is always a level of abstraction on which the reward is visible in a few steps~\cite{Kulkarni2016}.
  
\subsection{Exploration and the Influence Model} \label{s:exploration}

The previous sections described how the Expert is able to see into the future and decide what to predict in order to reach preferred states. However, because of the stochasticity of the environment (\cref{s:nonDeterministicAndNoisy}) and because the actions are not expressed explicitly (Appendix \ref{s:actions}), the Expert does not know how much power it has to influence the future states $\vec{x}(t+1), \ldots, \vec{x}(t+T_f)$ by sending the desired goal signal $Go(t)$. For this reason, it is not possible to use the passive model described in Appendix \ref{s:externalContext} which does not contain this information.

Instead, each Expert learns a separate \textit{influence model} which captures the conditional probability that a step $T_h + f, f \in 1 \isep T_f$ in a sequence $s_i = v_{i_1}, \ldots, v_{i_{T_h}}, v_{i_{T_h+1}} \ldots, v_{i_{m=T_h+T_f+1}}$ will happen $\left(\text{i.e., that } \vec{x}(t + f) = \delta(v_{i_{T_h + f +1}}) \right)$ given that all the previous steps $0, 1, \ldots, T_h + f - 1$ will have happened, the sequence of contexts will have been $\vec{\hat{c}} = \vec{\hat{c}}_1, \vec{\hat{c}}_2, \ldots, \vec{\hat{c}}_{T_h+f+1}$ and the Expert has actively tried to perform this step (i.e. is has predicted it on its Goal output $Go(t + f - 1) = \delta(v_{i_{T_h + f + 1}})$):   
\begin{multline} \label{e:influenceModel}
\Prior^I(f) = \Prior^I \left( \vec{x}(t + f) = \delta(v_{i_{T_h + f + 1}}) \vert \right.\\  
\vec{x}(t - T_h + d)) = \delta(v_{i_{d+1}}) \, \forall d = 0 \isep T_h + f - 1, \\
\vec{c}(t - T_h + e)) = \vec{\hat{c}}_e \, \forall e = 0 \isep T_h + f, \\
\left. Go(t+f-1)= \delta(v_{i_{Th+f+1}}) \right).
\end{multline}
It practice, we store the number of successful transitions observed during the agent's lifetime for each dimension of the context independently and then compute the resulting value $\Prior^I(f)$ for each $f \in 1 \isep T_f$ in the same way as in Appendix \ref{s:externalContext}. 

The probability $P^G(s_i)$ from the \cref{e:pre} is then computed as:
\begin{equation}
P^G(s_j) = P(s_j | \vec{x}(t-T_h:t)) \cdot \prod_{i = 1 \isep f} \Prior^I(f),
\end{equation}
where $P(s_j | \vec{x}(t-T_h:t))$ is computed as in \cref{e:sequenceProbNonnormalized} and $f$ is taken from \cref{e:eff}.

This influence model is then used in \cref{e:psi2} in all Experts in the situation when the agent is supposed to act.

However, because each Expert tries to maximize the reward in a greedy way (see Appendix \ref{s:rl}), its behavior would be biased towards the first rewards it found and the sequences leading potentially to higher rewards might be never explored. For this reason, it is necessary to add an explicit exploration mechanism which ensures that the influence model from \cref{e:influenceModel} is updated evenly for all sequences.  

The exploration can also utilize the fact that the learned model is represented in a hierarchical manner. By performing a random walk strategy (do a random action with exploration probability $\epsilon$) typically used in RL systems in a distributed manner, we obtain a powerful exploration mechanism. Performing a random step in an Expert high in the hierarchy means performing a whole complex policy\cite{Bacon2016,Hengst2000}, because such an Expert has a highly compressed and abstract representation of the world, where each cluster comprises of potentially multiple sequences of clusters on the level below, which themselves represent sequences of spatio-temporal representations from the lower layer, etc.     

As an example, children start learning actions by moving their own limbs mostly randomly. This can be seen as a low-level sensori-motoric hierarchy, where exploration is done in a space of primitive actions -- the moving of limbs. After some time, the child learns e.g. how to crawl. This new \textit{crawling from A to B} behavior can be at the same time seen as both 1) a complex policy (coordinated sequence of limb movements) and 2) a \qq{simple} action (move from A to B) on a higher level of the hierarchy. Now, the child can use this new \qq{simple} action to navigate randomly between rooms. Despite that the choice of target room can be purely random, this resulting hierarchical exploration of the environment is far more efficient than just the random movement of limbs.

\subsection{Unsupervised Learning of Disentangled Representations}\label{a:disentangled}

Going back to the passive mechanisms of the architecture, this section presents an optional mechanism which adds the ability to learn disentangled representations of observed data in an unsupervised way. It enables multiple Experts to efficiently decompose the input space into parts which are generated by independent hidden processes. Learning disentangled representations is vital for modelling the world in an efficient and compositional manner (as shown e.g. in~\cite{Higginsb,Thomas}).

Compared to recent models based on Deep Learning (DL) \cite{Higgins}, the presented approach is based on simple Experts and therefore is not so powerful. In this optional setting, multiple Experts can compete for the same observations by the mechanism inspired by predictive coding ~\cite{Spratling2017,Harpur1996}, but it can be also related to Dynamic Routing between Capsules~\cite{Sabour}. 

The Spatial Pooler of an Expert $i\in \mathcal{G}$ (a set of Experts called a \textit{predictive group}) receives a sequence of raw observations $\vec{o}^i(t)$ and learns the cluster centers $\mat{V}^i$. The output of the Spatial Pooler is then a one-hot vector determining the index of the closest learned pattern:
\begin{equation} \label{e:dist2}
    \vec{x}^i(t) = \vf{f}^i(\vec{o}^i(t)) = \argmin_{\vec{v}^i_j \in V^i} \left( \dist \left (\vec{v}^i_j, \vec{o}^i(t) \right ) \right).
\end{equation}
Here, the SP is used as an approximation of a generative model, therefore we also define a generative function, which takes the winning cluster $\vec{x}^i(t)$ and projects it into the input space:
\begin{equation} \label{e:inv}
	\vec{\hat{o}}^i(t) = \vf{f}'^i(\vec{x}^i(t)) = \vec{x}^i(t) \mat{V}^i,
\end{equation}
here $\vec{x}^i(t)$ is a one-hot $k$-dimensional row vector defining the currently winning cluster center (see \cref{e:dist2}), and $\mat{V}^i$ is a matrix containing one cluster center on each row. The multiplication of these results in the corresponding cluster center in the input space $\vec{\hat{o}}^i(t)$.

\begin{figure}[ht]
    \centering
    \includegraphics[width=0.9 \linewidth]{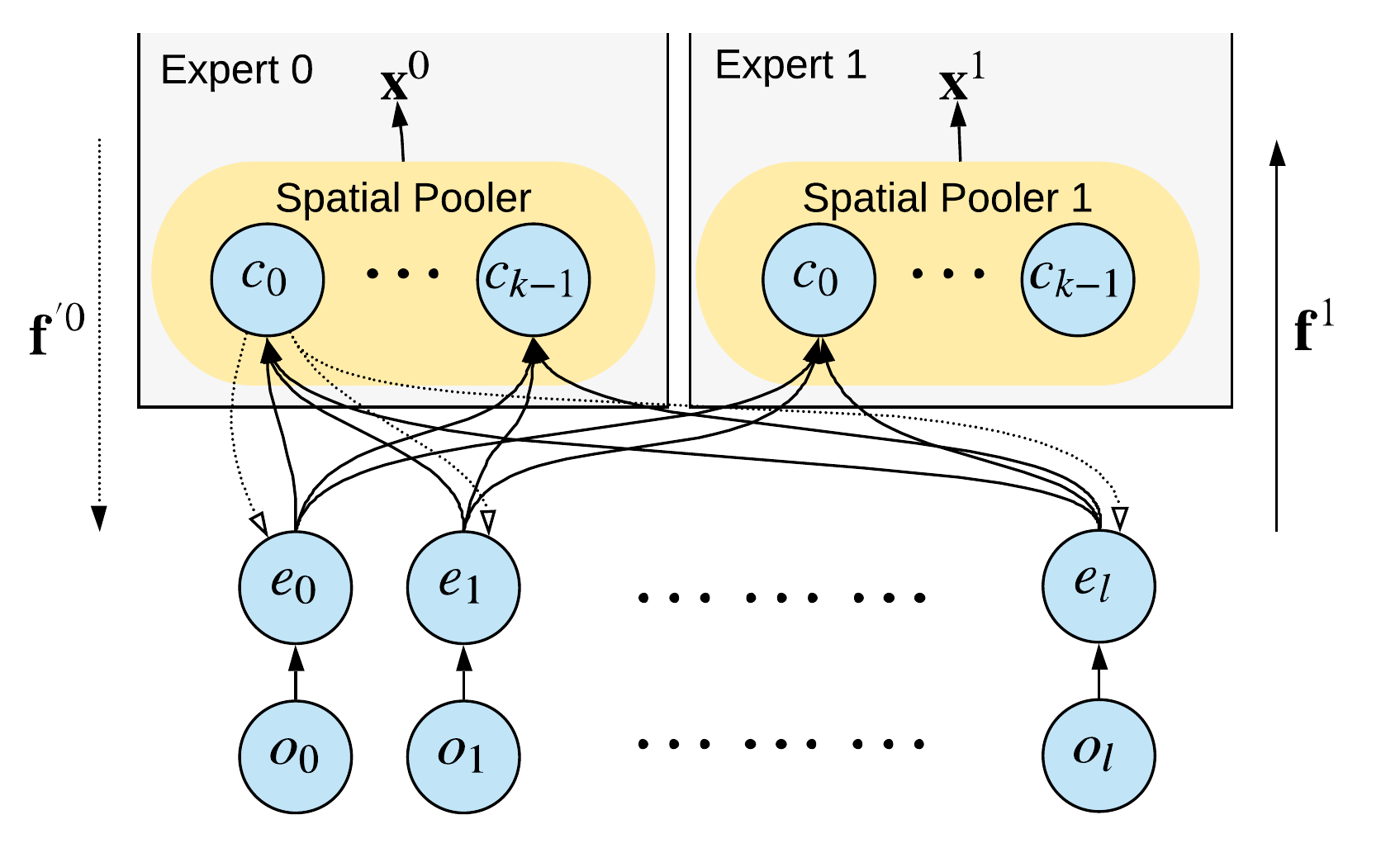}
    \caption{Illustration of the predictive group comparable to predictive coding algorithms. The Experts operate as usual (learning, inference), the forward/backward pass is just called iteratively for each observation.}
    \label{f:group}
\end{figure}

With this formalism, we can define the following two approximations of the predictive coding algorithms. One can think of the proposed algorithm as a group-restricted sparse coding version of them:

\subsubsection{Rao and Ballard approximation}

This approximation of Rao and Ballard's algorithm makes the interaction of the hidden units compatible with the Spatial Poolers. The interaction between Spatial Poolers of Experts is iterative, after the input is presented, the Spatial Poolers compete for the data as follows:
\begin{align}
	\vec{e}(t) &\leftarrow \vec{o}(t)- \sum_{i\in[1 \isep n]} \vec{\hat{o}}^i(t), \nonumber \\
	\vec{x}^i(t) &\leftarrow \vf{f} \left(\vec{\hat{o}}^i(t) + \mu \vec{e}(t) \right)
    \label{e:raoNew}
\end{align}
Similarly to the original equations shown in~\cite{Spratling2017}, the error vector $\vec{e}(t)$ is computed as the difference between the observation $\vec{o}(t)$ and the sum of the reconstructions $\vec{\hat{o}}^i(t)$ of all the Experts in the predictive group. This means that during the iteration, each Expert receives the part of the observation it is able to reconstruct $\vec{\hat{o}}^i(t)$, plus the overall residual error $\vec{e}(t)$. Note that in this notation, the $t$ denotes the time step of the input observation $\vec{o}(t)$, the architecture can perform multiple iterations during one time step.

\subsubsection{PC/BC-DIM approximation}

The second method of disentangling the representations is based on an approximation of the Predictive Coding/Biased Competition-Divisive Input Modulation (PC/BC-DIM) algorithm:
\begin{align}
	\vec{e}(t) &\leftarrow \vec{o}(t) \eldiv \left(\vecg{\epsilon}_2 + \sum_{i\in[1 \isep n]} \vec{\hat{o}}^i(t)\right), \nonumber \\ 
	\vec{x}^i(t) &\leftarrow \vf{f} \left( \left( \vecg{\epsilon}_1 + \vec{\hat{o}}^i(t) \right) \elmult \vec{e}(t) \right).
    \label{e:pcbcNew}
\end{align}
Note that the original version (shown in~\cite{Spratling2017}) encourages an increase  in activity of those hidden neurons $\vec{y}$ which are both active and able to mitigate the residual error $\vec{e}$ well. Compared to this, our version computes the element-wise product in the original input space $\vec{o}(t)$. This means that each Expert receives higher values at positions in $\vec{o}(t)$ which it already reconstructs and which contain some residual error. The first part of the equation remains similar to the original version: it amplifies parts of the input which are not yet reconstructed well.

The resulting mechanism enables the experts to represent the observations in a compositional way. It was experimentally shown that in the case where the input is generated by $N$ independent latent factors and $N$ Experts are used, the architecture is able to represent each factor in one Expert. Compared to this, in case the input is generated by just $M$ latent factors, where $M<N$ the group of $N$ Experts will form a \textbf{sparse distributed representation} of the input.

\end{document}